\documentclass[10pt,twocolumn,letterpaper]{article}

\usepackage{cvpr}              % To produce the CAMERA-READY version
\usepackage{graphicx}
\usepackage{amsmath}
\usepackage{amssymb}
\usepackage{booktabs}
\usepackage{bm} 
\usepackage{multirow}
\usepackage{listings}
\usepackage{xcolor}
\usepackage[table]{xcolor}
\newcommand{\cmark}{\checkmark}
\newcommand{\colorpatchframe}[3]{\raisebox{-0.5ex}{{\setlength{\fboxsep}{0pt}\fcolorbox{gray}{white}{\color[RGB]{#1,#2,#3}\rule{2.2ex}{2.2ex}}}}}

\renewcommand{\textcolor}[2]{#2}

\definecolor{cvprblue}{rgb}{0.21,0.49,0.74}
\usepackage[pagebackref,breaklinks,colorlinks,allcolors=cvprblue]{hyperref}

\def\paperID{\#\#\#\#} % *** Enter the Paper ID here
\def\confName{CVPR}
\def\confYear{2026}

\title{\raisebox{-0.2\height}{\includegraphics[height=0.75cm]{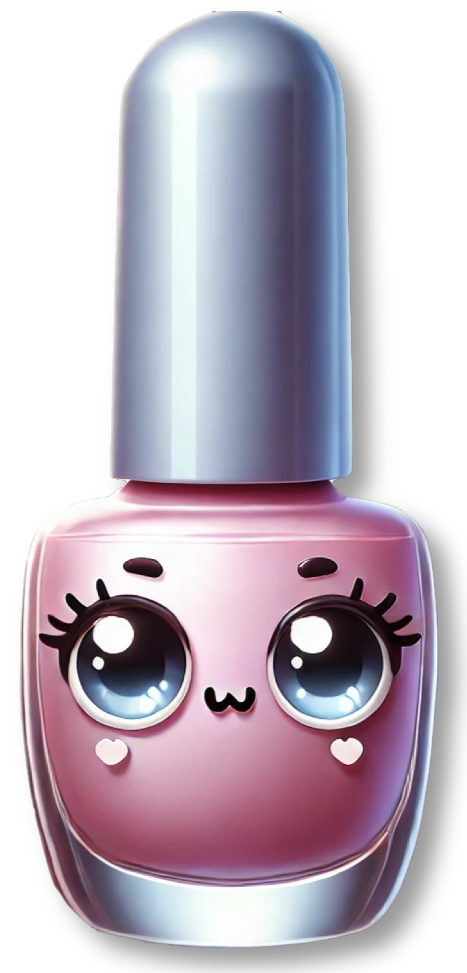}}\hspace{0.2em}%
NaiLIA: Multimodal Nail Design Retrieval\\ Based on Dense Intent Descriptions and Palette Queries}

\author{
Kanon Amemiya \quad Daichi Yashima \quad Kei Katsumata \quad Takumi Komatsu\\
Ryosuke Korekata \quad Seitaro Otsuki \quad Komei Sugiura\\
Keio University
}

\begin{document}
\maketitle
\vspace{-6mm}
\begin{abstract}
We focus on the task of retrieving nail design images based on dense intent descriptions, which represent multi-layered user intent for nail designs.
This is challenging because such descriptions specify unconstrained painted elements and pre-manufactured embellishments as well as visual characteristics, themes, and overall impressions.
\textcolor{blue}{In addition to these descriptions, we assume that users provide palette queries by specifying zero or more colors via a color picker, enabling the expression of subtle and continuous color nuances.}
Existing vision-language foundation models often struggle to incorporate such descriptions \textcolor{blue}{and palettes}.
To address this, we propose NaiLIA, a multimodal retrieval method for nail design images, which comprehensively aligns with dense intent descriptions \textcolor{blue}{and palette queries} during retrieval.
Our approach introduces a relaxed loss based on confidence scores for unlabeled images that can align with the descriptions.
To evaluate NaiLIA, we constructed a benchmark consisting of 10,625 images collected from people with diverse cultural backgrounds.
The images were annotated with long and dense intent descriptions given by over 200 annotators.
Experimental results demonstrate that NaiLIA outperforms standard methods.
% Our project page, code, and datasets are provided in the supplementary materials.
Our project page, code, and dataset are available at: \href{https://nailia-94dpr.kinsta.page/}{https://nailia-94dpr.kinsta.page/}.
\end{abstract}
\vspace{-4mm}
\section{Introduction}
\vspace{-2mm}

% \begin{figure}[t]
%     \centering
%     \includegraphics[width=\linewidth]{fig/eye-catch.png}
%     \caption{A typical use case for our task. The user inputs a dense intent description, expressing their detailed and multi-layered intent for nail designs to a nail artist. NaiLIA retrieves images that align with the description. In this example, the top-ranked nail design is painted with expressive creativity (``fins and shells with a light blue appearance''), and the shells are decorated with pre-manufactured embellishments (``pearl nail accessories''). These elements represent a theme (``mermaid''), creating an overall impression (``a fresh and sparkling look'').}
%     \label{fig:eye-catch}
% \end{figure}

\begin{figure}[t]
    \centering
    \vspace{-3mm}
    \includegraphics[width=\linewidth]{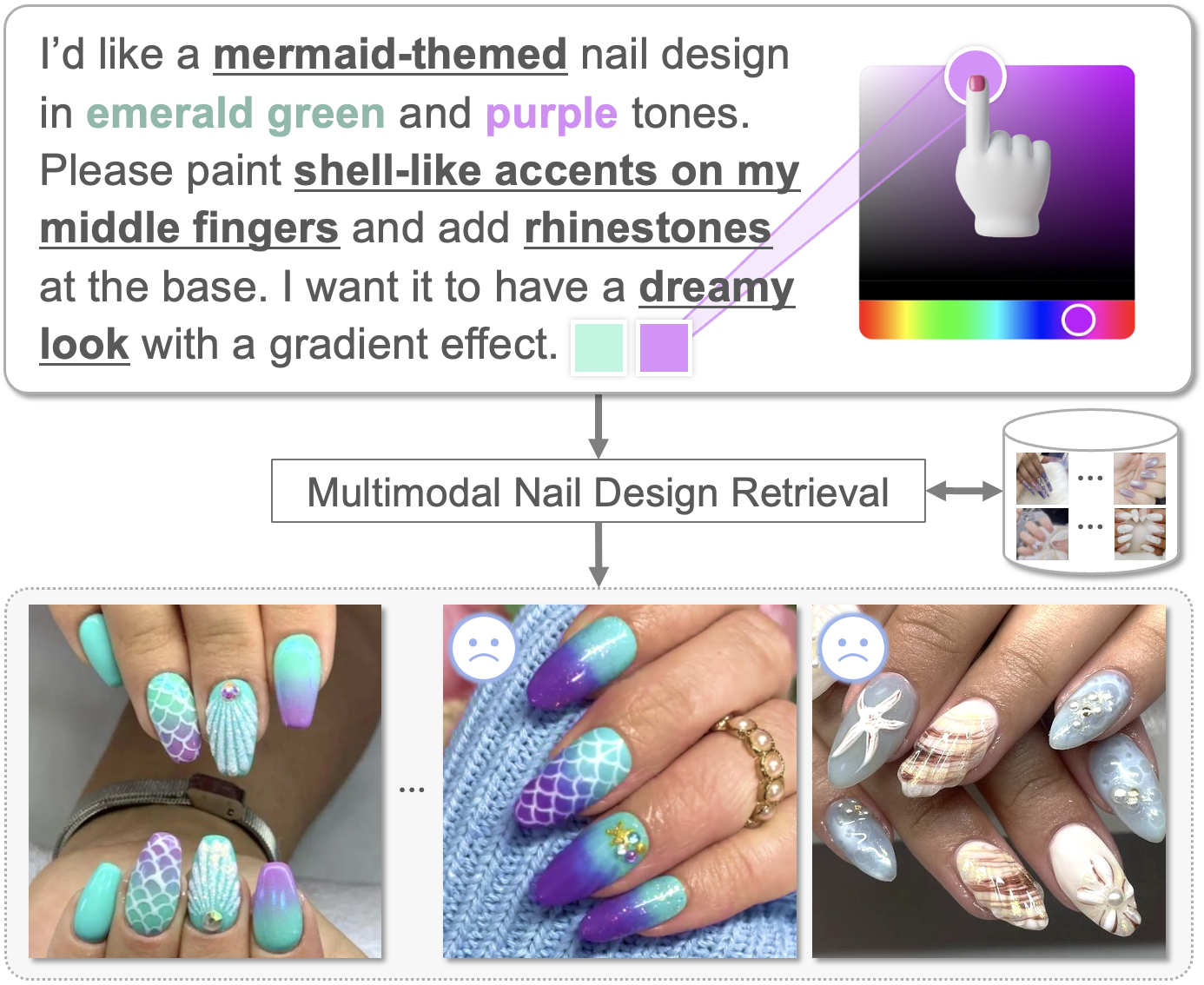}
    \caption{A typical use case for our task. \textcolor{blue}{The user inputs a dense intent description and an optional palette query. The palette query allows the user to select zero or more colors through a color picker interface. The model should rank the leftmost image higher than the middle image (painted in a darker purple than specified by the palette query) and the rightmost image (featuring realistic shell ornaments rather than the intended shell-inspired design).} Details of this task are provided in \Cref{sec:statement}.}
    \vspace{-5mm}
    \label{fig:eye-catch}
\end{figure}

% 1段落
The global nail salon market is valued at approximately USD 11 billion, indicating a significant demand for the ability to search for nail designs meeting user preferences and/or nail artists capable of applying these designs.
When customers request nail designs from nail artists, they often articulate their preferences \textcolor{blue}{by specifying painted elements (e.g., patterns) and decorative elements (e.g., nail accessories), as well as the overall themes and impressions that emerge from the interplay of these elements.}
It would be beneficial if the customers could search for nail design images using such \textbf{\textit{dense intent descriptions}}. 

\textcolor{blue}{Although one can imagine that a simple text-to-image approach would suffice, retrieval-based approaches using real-world images are more practical. Indeed, nail artists have reported difficulties when clients present AI-generated references because such images often violate physical constraints (e.g., unrealizable nail accessories), making faithful reproduction difficult or infeasible. Moreover, several beauty service platforms have begun restricting the use or posting of AI-generated images.}

% 3段落
\textcolor{blue}{
In this study, we focus on the task of retrieving nail design images based on dense intent descriptions. 
The descriptions represent user intent for nail designs in a detailed and multi-layered manner (e.g., colors, patterns, themes, and overall impressions). 
We assume that users can optionally specify colors (\textbf{\textit{a palette query}}) via a color picker, enabling the expression of subtle and continuous color nuances.
% The palette queries enable users to continuously specify multiple colors via color pickers, allowing the expression of continuous color nuances that cannot be represented solely through discretized linguistic descriptions.
\cref{fig:eye-catch} shows a typical use case for our task.
Given the description \textit{``I’d like a mermaid-themed, ... I want it to have a dreamy look with a gradient effect,''} and the palette query consisting of \colorpatchframe{178}{247}{225} (\#b2f7e1) and \colorpatchframe{222}{143}{252} (\#de8ffc), the model should retrieve the leftmost image at a high rank.}

% 2段落
Retrieving nail designs that closely align with dense intent descriptions is very challenging.
This difficulty arises mainly because a nail design generally consists of a painted portion, which allows for creative flexibility, and a decorative portion, which can only be modified through the selection and arrangement of pre-manufactured embellishments. 
Furthermore, the descriptions often include themes and spatial relationships of the designs, in addition to visual characteristics.
This is difficult because a theme (e.g., ``a mermaid-themed'') represents an abstract design concept rather than a concrete entity (e.g., an actual mermaid).
% Moreover, in instances where a description specifies a design for a particular finger (e.g., ``a seashell on the ring finger''), it is necessary to capture the correspondence between each finger and the design in the image. 

% 4段落
Although existing vision-language foundation models \cite{radford2021learning, Wang_2023_CVPR, 10377550} have achieved promising results on related tasks, applying them directly to this task does not always yield satisfactory results.
Specifically, they tend to have an inherent bias toward retrieving nail designs at a specific level of abstraction, which is often photorealistic \textcolor{blue}{(\cref{fig:eye-catch}, the shells on the rightmost nail design)}.
This is primarily because they rely on the InfoNCE loss \cite{oord2018representation}, which treats all non-positive samples as negatives.
% This is insufficient because the phrase ``flower nail parts'' could refer to uncolored metallic parts shaped like flower silhouettes or character-based mascot parts inspired by flowers. 
% Furthermore, the vision-language foundation models often fail to incorporate multi-layered intent descriptions. 
% Furthermore, they often fail to incorporate multi-layered intent descriptions. 
% Instead, they tend to prioritize nail designs that align with only specific aspects of the descriptions.
\textcolor{blue}{Furthermore, while subtle color nuances are critical in fashion domains, continuous color inputs (e.g., palette queries) have been largely overlooked.
As a result, even when a user intends a color such as \colorpatchframe{222}{143}{252} (\#de8ffc), the models may rank undesired, darker-toned designs higher (\cref{fig:eye-catch}, the middle nail design).}

% 5段落
To address these issues, we propose \textbf{NaiLIA}, a multimodal retrieval model for nail design images based on dense intent descriptions \textcolor{blue}{and palette queries}.
% This method enables the retrieval of nail design images that comprehensively align with descriptions in which dense user intent is expressed in a detailed and multi-layered manner.
% NaiLIA differs from existing approaches in the following aspects.
% First, 
NaiLIA estimates confidence scores for images that align with the given description and can be considered positive examples but are not explicitly labeled (\textbf{\textit{unlabeled positives}}).
% \textcolor{blue}{Incorporating the scores into the loss function further enhances training  efficiency, as it avoids undesired anti-correlation between pairs that should  be correlated.
% In addition, NaiLIA models the relationship between dense intent descriptions and palette queries, enabling it to rank nail designs that closely align with the user's intended color tones more highly.}
% It then incorporates the scores into a loss function.
% This approach enables effective learning because there are numerous unlabeled positives in this task.
% This approach increases training efficiency by avoiding undesired anti-correlation between pairs that should be correlated.
% Second, the proposed method decomposes the dense intent descriptions and structures natural language descriptions of nail design images to align them in a multi-layered manner.
% Second, the proposed method incorporates structured natural language descriptions of nail design images with visual features extracted through both unimodal and multimodal encoders.
% Furthermore, to handle the multi-layered nature of user intent, we structure the descriptions across four distinct design levels: nail shape, primitive visual characteristics of paints and decorations, themes, and impressions.
% These approaches capture design concepts and complex referring expressions, as well as the detailed intent.

The main contributions of this study are:
\begin{itemize}
    \setlength{\parskip}{0.2mm} % 段落間
    \setlength{\itemsep}{0.2mm} % 項目間
    \item We introduce the Confidence-based Relaxed Alignment Module, which estimates confidence scores of unlabeled positives and incorporates the scores into a relaxed loss. 
    \item \textcolor{blue}{To capture the multi-layered user intent in dense intent descriptions and the continuous multi-color preferences expressed in palette queries, we introduce the Intent-Palette Fusion Module, which models the relationship between descriptions and palette queries.}
    \item We introduce the Visual Design Fusion Module, which integrates three types of visual representations: (i) a unimodal visual representation, (ii) a multimodal aligned representation, and (iii) an img2txt intent structure representation, which includes design concepts and spatial relations through natural language.
    \item \textcolor{blue}{We construct the NAIL-STAR benchmark, which consists of pairs of a dense intent description and a palette query, and their corresponding diverse nail design images.}
\end{itemize}
\section{Related Work}
\vspace{-1mm}
%2.1
Numerous studies on fashion AI have been conducted in various fields, including retrieval~\cite{Park_2019_CVPR_Workshops, Kuang_2019_ICCV}, generation~\cite{Han_2018_CVPR, choi2021viton, Ning_2024_CVPR, Yu_2019_ICCV}, trend forecasting~\cite{Al-Halah_2017_ICCV}, and compatibility evaluation~\cite{Pang_2024_WACV}.
Several survey papers ~\cite{Imtiaz_Pathirana_Saheel_Karunanayaka_Trenado_2024, kouslis2024ai} provide a comprehensive overview of fashion AI, including methods, standard datasets, and evaluation metrics.
% In addition, multimodal retrieval has been applied in diverse fields such as medical diagnosis~\cite{vanSonsbeek2023xtra, zhang2022deep} and robotics~\cite{10816380, korekata2025dm2rm}.  
% Recent advances in multimodal retrieval methods have been significantly driven by vision--language foundation models such as CLIP~\cite{radford2021learning} and BEiT-3~\cite{Wang_2023_CVPR}.

\vspace{-3mm}
\paragraph{Multimodal retrieval in fashion AI.}
Multimodal retrieval has been applied in diverse fields such as medical diagnosis~\cite{vanSonsbeek2023xtra, zhang2022deep} and robotics~\cite{relaxformer, korekata2025dm2rm}.  
%2.3
Extensive research has been conducted on multimodal retrieval within the fashion domain, encompassing tasks such as cross-modal retrieval (XMR)~\cite{9878892, rostamzadeh2018fashion, FashionBERT} and text-guided image retrieval (TGIR)~\cite{10484044, 10568424, Lee_2021_CVPR}.
XMR aims to retrieve the most relevant image or text from a collection based on a given text or image query.
% Compared to the general vision--language domain, models tackling fashion-based tasks require task-specific knowledge, such as color, shape, and texture to capture the subtle distinctions of style and design.
% Notable methods include FashionBERT~\cite{FashionBERT}, Kaleido-BERT~\cite{Zhuge_2021_CVPR}, and EI-CLIP~\cite{9878892}, which extend foundational models to focus on aspects such as design and style.
% EI-CLIP extends CLIP by incorporating fashion-specific terminology and metadata selection, including visual attributes, and further introduces an intervention-based, entity-aware contrastive learning framework.
In contrast to models in the general vision--language domain, models tackling fashion-based tasks~\cite{FashionBERT, Zhuge_2021_CVPR} require task-specific knowledge, such as color, shape, and texture to capture the subtle distinctions of style and design.
For example, EI-CLIP~\cite{9878892} extends CLIP by incorporating fashion-specific terminology and metadata selection, and it further introduces an intervention-based, entity-aware contrastive learning framework.
By contrast, another line of research~\cite{10484044, 10568424, Lee_2021_CVPR} has tackled TGIR using a multimodal query, consisting of a reference image and a modifying text, to search a collection of images.
CoSMo~\cite{Lee_2021_CVPR} processes the content and style of the reference image independently under the guidance of textual instructions.
\textcolor{blue}{
Despite this progress, few multimodal retrieval methods are designed to handle multi-color inputs represented by continuous values (e.g., RGB).}
% In contrast, NaiLIA processes palette queries composed of multiple colors, in addition to textual descriptions, enabling the retrieval of images that better align with the user's desired colors.

\begin{figure*}[t]
    \centering
    \includegraphics[width=0.98\linewidth]{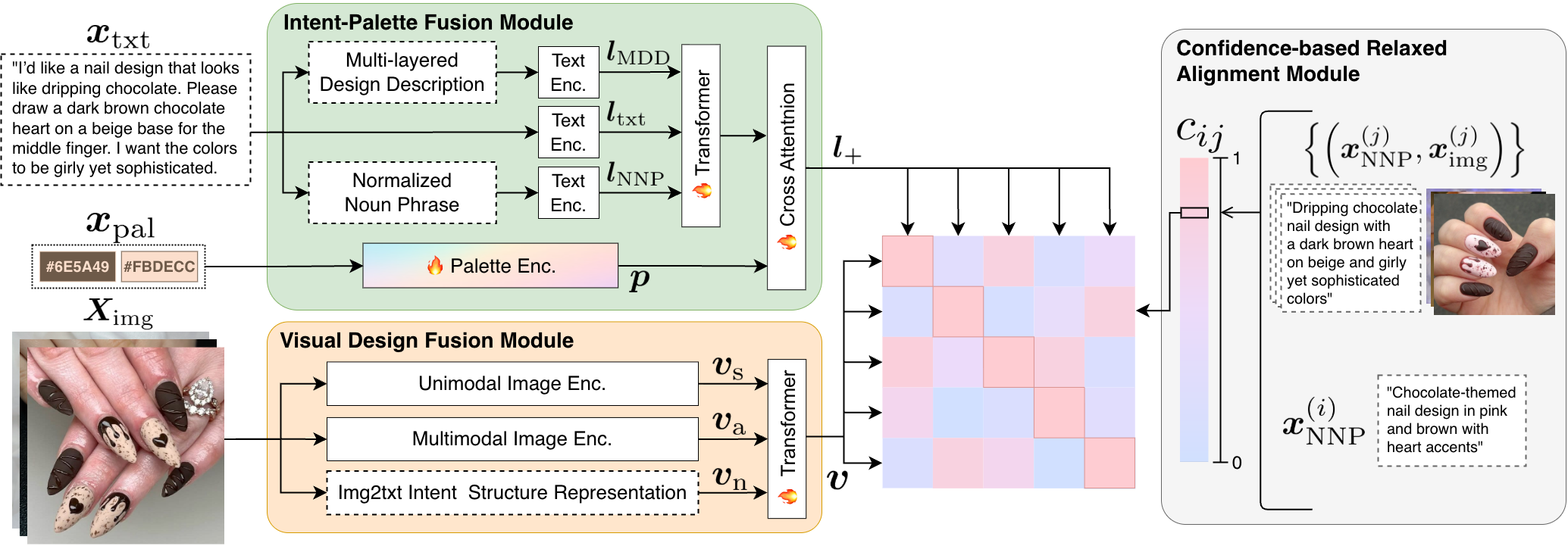}
    \vspace{-2mm}
    \caption{Architecture of NaiLIA. \textcolor{blue}{A language-palette representation $\bm{l}_\text{+}$ is extracted from $\bm{x}_\text{txt}$ and $\bm{x}_\text{pal}$ by the Intent-Palette Fusion Module}, and a visual representation $\bm{v}$ is extracted from $\bm{x}_\text{img}$ by the Visual Design Fusion Module. In the Confidence-based Relaxed Alignment Module, unlabeled positives are assigned, and the loss is calculated considering the unlabeled positives. \textcolor{blue}{Here, Enc. denotes the encoder.}}
    \vspace{-3mm}
    \label{fig:model}
\end{figure*}   

\vspace{-3mm}
\paragraph{Vision--language foundation models.}
%2.4
In multimodal retrieval tasks, many vision--language foundation models (e.g., \cite{pmlr-v202-li23q, Wang_2023_CVPR}) have achieved promising performance. 
These foundation models can be applied to a wide range of downstream tasks, including visual question answering and image captioning, in addition to retrieval tasks.
Furthermore, extensive research has been conducted on the extension of CLIP~\cite{Sun_2024_CVPR, zhang2025long, 10377550, llm2vec}.
For instance, AlphaCLIP~\cite{Sun_2024_CVPR} introduced an alpha channel to identify regions of interest, allowing the model to focus on specific areas while maintaining CLIP's visual recognition accuracy. 
However, for certain tasks, the image encoder of a vision--language foundation model alone might not have sufficient capacity to represent visual features or could introduce biases~\cite{ kar2025brave}.
Several studies~\cite{goko2024task, kar2025brave, SPHINX} have shown that integrating the visual features of multiple foundation models by employing unimodal image encoders in addition to the image encoder in vision--language foundation models can achieve promising results in the vision--language tasks.
\textcolor{blue}{In contrast to prior studies, NaiLIA does not merely fuse features extracted from multiple encoders, as explained in \Cref{sec:vdfm}.}
% By introducing a visual representation mediated through natural language, it enables three heterogeneous visual representations to complement and reinforce one another.

% Contrastive learning is widely adopted in vision--language foundation models (e.g., \cite{radford2021learning, li2022blip}). 
% CLIP employs InfoNCE~\cite{oord2018representation} as its contrastive loss function. 
% However, training with InfoNCE assumes a strict one-to-one correspondence between a single text and a single image as a positive example. 
% Consequently, samples that are semantically aligned but not explicitly labeled as positive pairs are treated as negatives, introducing noise into the learning process~\cite{wu2022data}.
% To address this issue, various loss functions have been proposed to relax the one-to-one labeling constraint in InfoNCE-based training (e.g., \cite{wu2022data, Lin_2023_WACV, 10816380, gao2022pyramidclip}).
% ReCo~\cite{Lin_2023_WACV} mitigates this issue by relaxing the constraints on negative pairs, ignoring those with similarity scores below zero, it does not explicitly handle unlabeled positives.
\vspace{-3mm}
\paragraph{Benchmarks for multimodal retrieval.}
%2.6
% Representative benchmarks for fashion AI research include Fashion-Gen~\cite{rostamzadeh2018fashion}, Fashion200K~\cite{Han_2017_ICCV}, and FashionIQ~\cite{Wu_2021_CVPR} for multimodal retrieval, as well as VITON-HD~\cite{choi2021viton} and Makeup Transfer dataset~\cite{li2018beautygan} for image synthesis.
Several benchmarks exist for multimodal retrieval in fashion AI~\cite{rostamzadeh2018fashion, Han_2017_ICCV, Wu_2021_CVPR}.
FashionIQ~\cite{Wu_2021_CVPR} is used to evaluate interactive fashion image retrieval and incorporates natural language feedback by including relative captions describing the differences in the attributes and styles of two fashion images.
Fashion-Gen~\cite{rostamzadeh2018fashion} serves as a benchmark for multiple vision--language tasks on fashion images, including fashion image captioning, category recognition, and fashion image generation.
For multimodal retrieval tasks, various benchmarks~\cite{lin2014microsoft, Plummer_2015_ICCV} beyond the fashion domain exist.
Most of them use very simple and short phrases; however, this is insufficient for our task.

\vspace{-1mm}
\section{Method}
\vspace{-1mm}
\subsection{Problem Statement}
\vspace{-1mm}
\label{sec:statement}
%3.1
We define the \textbf{Nail} design \textbf{S}emantic \textbf{T}ext-\textcolor{blue}{palette} \textbf{A}ligned \textbf{R}etrieval (\textbf{NAIL-STAR}) task as follows: \textcolor{blue}{given a dense intent description and a palette query for a nail design, the goal is to retrieve a ranked list of appropriate images.}
% Nail designs are often composed of a painted portion that allows for creative flexibility, as well as a decorative portion that can be modified only by selecting and arranging pre-manufactured embellishments.
% The description typically includes not only visual information but also the themes expressed through the balance between the painted and decorative portions, and the impressions they create.
%3.3
% \cref{fig:eye-catch} illustrates a typical use case for the NAIL-STAR task.
% \textcolor{blue}{Given the input description and palette query, the model should retrieve the leftmost image.
\textcolor{blue}{
As illustrated in the use case in \cref{fig:eye-catch}, the leftmost image should be retrieved at a high rank.
This design is painted with expressive creativity (``mermaid fin patterns and shell''), and the shells are decorated with pre-manufactured embellishments (``rhinestones'').
These elements symbolize a theme (``mermaid''), forming an overall impression (``dreamy look'').
Moreover, the design's color tones closely align with \colorpatchframe{178}{247}{225} (\#b2f7e1) and \colorpatchframe{222}{143}{252} (\#de8ffc).}

The terms used in this study are defined as follows.
A \textbf{dense intent description} refers to a text that represents user intent for nail design in a detailed and multi-layered manner. 
\textcolor{blue}{A \textbf{palette query} consists of zero or more colors specified by the user through color pickers to define the desired color scheme.}
A \textbf{target nail design image} is an image that is labeled as a positive.
An \textbf{unlabeled positive} denotes an image that could be considered a target nail design image but lacks explicit labeling.

%3.7
% In this study, we focus on images featuring fingernails that have been styled with painted elements, decorations, or fake nails.
We focus on image retrieval rather than image synthesis.
While a text-to-image system for generating nail design images can be envisioned, the prevalence of nonexistent embellishments in the generated output renders such a system impractical, in contrast to a search-based system.
Furthermore, we assume a user-agnostic retrieval setting and do not incorporate user-specific personalization.
%3.8
% We use the mean reciprocal rank (MRR) and recall@$K (K=1,5,10,20)$ as evaluation metrics, with recall@10 as the primary metric.

\subsection{Overview of the Proposed Method}
%4.1
We propose \textbf{NaiLIA}, a multimodal retrieval method for nail design images based on dense intent descriptions \textcolor{blue}{and palette queries}.
% 4.5
\cref{fig:model} shows the structure of NaiLIA.
% 4.4
It consists of three modules: Intent-Palette Fusion Module (IPFM), Visual Design Fusion Module (VDFM), and Confidence-based Relaxed Alignment Module (CRAM).
%4.2
NaiLIA estimates the confidence scores of unlabeled positives and incorporates these scores into the contrastive loss.
This approach can be broadly applied to other multimodal retrieval tasks that involve numerous similar images~\cite{Wu_2021_CVPR, 8099810}.
%4.3
% The novelties of our method are as follows:
% \begin{itemize}
%     \setlength{\parskip}{0.2mm} % 段落間
%     \setlength{\itemsep}{0.2mm} % 項目間
%     \item We introduce the CRAM, which estimates confidence scores of unlabeled positives and incorporates these scores into the loss function to enable model training with unlabeled positives. 
%     \item We introduce the VDFM, which integrates three types of visual representations: (i) unimodal visual representation, (ii) multimodal aligned representation, and (iii) img2txt intent structure representation that includes design concepts, complex referring relations, and other intricate aspects through natural language.
%     \item To capture the multi-layered user intent in dense intent descriptions, we introduce the LIFM that structurally models the normalized descriptions in terms of dense intent.
% \end{itemize}

%4.6
We define the input $\bm{x}$ to the model as follows:
\textcolor{blue}{
$
\bm{x} = \{ \bm{x}_{\text{txt}},  \bm{x}_{\text{pal}}, X_{\text{img}}\}, 
X_{\text{img}} = \{\bm{x}_{\text{img}}^{(i)} \mid i = 1, \dots, N_{\text{img}}\},
$
where $\bm{x}_{\text{txt}}$, ${\bm{x}_{\text{pal}} \in \mathbb{R}^{3 \times N_{\text{pal}}}}$ and $\bm{x}_{\text{img}}^{(i)} \in \mathbb{R}^{3 \times W \times H}$ denote a description, a palette query, and the $i$-th nail design image, respectively.
Here, $i$, $N_{\text{img}}$, $N_{\text{pal}}$, $W$, and $H$ denote the index of each image, number of images to be ranked, number of colors in the palette query, width of the image, and height of the image, respectively.}

\subsection{Intent-Palette Fusion Module}
\label{sec:ipfm}
\textcolor{blue}{
Given $\bm{x}_\text{txt}$ and $\bm{x}_\text{pal}$ as input, IPFM structures $\bm{x}_\text{txt}$ and models the relationship between the structures and $\bm{x}_\text{pal}$.
As illustrated in \cref{fig:eye-catch}, typical descriptions include elements at different levels of abstraction, leading to a lack of clarity and causing redundancy. 
To address this, we generate \textit{a multi-layered design description} derived from $\bm{x}_\text{txt}$ and structured into distinct components by leveraging an LLM.
In addition, to highlight key design aspects, we obtain \textit{a normalized noun phrase} that condenses the information in $\bm{x}_\text{txt}$.
Appendices~\ref{sec:prompt} and \ref{sec:discussion_hallucinations} provide the example prompts with generated texts and the discussion on possible LLM hallucinations, respectively.
}
\textcolor{blue}{
We then obtain language representations $\left( \bm{l}_\text{txt}, \bm{l}_\text{MDD}, \bm{l}_\text{NNP} \right)$, which are derived from $\bm{x}_\text{txt}$, the multi-layered design description, and the normalized noun phrase, respectively. 
We use multiple text encoders (e.g., \cite{Wang_2023_CVPR, 10377550, stella_en_400M_v5_2024}) to obtain $\left( \bm{l}_\text{txt}, \bm{l}_\text{MDD}, \bm{l}_\text{NNP} \right)$.
% Here, ${d_{\text{txt}}}$ denotes the output dimensions of the encoders, with the detailed settings provided in Appendix~\ref{sec:implementation}.
}
% We integrate these intent representations as follows:
% {\setlength{\abovedisplayskip}{4pt}
%  \setlength{\belowdisplayskip}{4pt}
% \begin{align}
% \bm{l} = \text{Transformer} \left( \left[ \bm{l}_{\text{txt}}; \bm{l}_{\text{mdd}}; \bm{l}_{\text{nnp}} \right] \right),
% \end{align}}
% where $\text{Transformer}(\cdot)$ denotes a Transformer encoder. 
% In parallel, we obtain the palette representation $\bm{p}\in \mathbb{R}^{d_{\text{txt}}}$ by feeding $\bm{x}_\text{pal}$ into a Transformer-based palette encoder as follows:
% {\setlength{\abovedisplayskip}{4pt}
%  \setlength{\belowdisplayskip}{4pt}
% \begin{align}
% \bm{p} = \text{Transformer} \left( f_\text{lab}(\bm{x}_\text{pal}) \right),
% \end{align}}
% where $f_\text{lab}$ represents the conversion from RGB values to the Lab color space, which is perceptually aligned with human vision.

\textcolor{blue}{
Motivated by the widespread use of palette queries in e-commerce systems as a natural input modality, we explicitly integrate them into our model design.
We obtain the palette representation $\bm{p}$ by feeding $\bm{x}_\text{pal}$ into a palette encoder that converts the color values from the RGB to the CIELAB color space and applies Transformer layers.
We then calculate attention scores over $\left( \bm{l}_\text{txt}, \bm{l}_\text{MDD}, \bm{l}_\text{NNP} \right)$ based on $\bm{p}$,  selectively emphasizing color-related elements. This is motivated by the presence of explicit color terms and expressions that implicitly indicate color tones in $\bm{x}_\text{txt}$.
Thus, a language-palette representation $\bm{l}_\text{+}$ is computed as follows:
{\setlength{\abovedisplayskip}{4pt}
 \setlength{\belowdisplayskip}{4pt}
\begin{align}
\bm{l}_\text{+} = \text{CrossAttn} (\bm{p}, \text{TFLayers} \left( \left[ \bm{l}_{\text{txt}}; \bm{l}_{\text{MDD}}; \bm{l}_{\text{NNP}} \right] \right)),
\end{align}}%
where $\text{CrossAttn}(\cdot,\cdot)$ and $\text{TFLayers}(\cdot)$ denote a cross-attention block and Transformer layers, respectively.
}

\subsection{Visual Design Fusion Module}
\vspace{-1mm}
\label{sec:vdfm}
%4.9
% The VDFM takes $\bm{x}_{\text{img}}^{(i)}$ as input.
% This module integrates three types of visual representations from nail design images: (i) a unimodal visual representation, (ii) a multimodal aligned representation, and (iii) an img2txt intent structure representation that handles design concepts and spatial relations.
% For example, the nail design prominently displayed on the bottom-left corner of \cref{fig:eye-catch} features a mermaid theme, with a fin on the middle fingers and a seashell on the ring fingers.
% We use this architecture because using only unimodal image encoders on the image does not sufficiently capture its abstract thematic elements and the correspondence between individual fingers and their specific designs, as shown in \Cref{sec:qualitative}.
% To overcome this limitation, we obtain comprehensive representations by integrating representation extracted from unimodal image encoders and multimodal image encoders, as well as representation derived from explanations generated by MLLMs.
\textcolor{blue}{
While some prior methods~\cite{SPHINX, kar2025brave} fuse unimodal and multimodal image encoders, they sometimes struggle to capture abstract concepts (e.g., character-based mascot inspired by flowers). 
% Furthermore, others~\cite{10655193} integrate features from unimodal encoders and LLMs, their alignment capabilities remain limited. 
% In particular, they struggle to align fine-grained visual attributes such as subtle differences in color and texture with corresponding textual descriptions.
% In contrast, the VDFM models the relationships among (i) visual features for colors and shapes, (ii) features aligned with natural language, and (iii) visual representations that include abstract concepts through natural language. 
% This is inherently effective and broadly applicable to retrieval tasks that require understanding of both abstract and fine-grained visual semantics.
To address this, VDFM introduces the following three representations:
}
% This module integrates the following three types of visual representations from nail design images.

\vspace{-4mm}
\paragraph{Unimodal visual representation.}
VDFM takes $\bm{x}_{\text{img}}^{(i)}$ as input.
We obtain unimodal visual representation $\bm{v}_{\text{s}}^{(i)}$ from $\bm{x}_{\text{img}}^{(i)}$, utilizing unimodal image encoders (e.g., \cite{dosovitskiy2020image, Jose2024DINOv2MT}) to capture visual characteristics such as the colors, shapes, and textures of the nail design.
% Here, ${d_{\text{s}}}$ denotes the output dimensions of the unimodal image encoders.

\vspace{-4mm}
\paragraph{Multimodal aligned representation.}
We use multimodal image encoders (e.g., \cite{Wang_2023_CVPR, 10377550}) to extract multimodal aligned representation $\bm{v}_{\text{a}}^{(i)}$ from $\bm{x}_{\text{img}}^{(i)}$. This obtains a representation aligned with language.
% Here, ${d_{\text{a}}}$ denotes the output dimensions of the multimodal image encoders.

\vspace{-4mm}
\paragraph{Img2txt intent structure representation.}
We employ multiple MLLMs to generate explanations for $\bm{x}_{\text{img}}^{(i)}$ in terms of painted elements, decoration, themes, and impressions of the design.
% When $\bm{x}_{\text{img}}^{(i)}$ contains distinctive patterns or nail accessories on specific fingers, the explanations also specify the relationships between finger names and the design elements.
Although employing a single MLLM might be sufficient, leveraging multiple MLLMs can mitigate the adverse effects of hallucinations potentially arising from individual MLLMs.
% The supplementary material includes prompts and examples of the explanations generated from them.
Appendix \ref{sec:prompt} includes prompts and examples of the explanations generated from them.

Subsequently, we use text encoders (e.g., \cite{Wang_2023_CVPR, 10377550, stella_en_400M_v5_2024}) to derive img2txt intent structure representation $\bm{v}_{\text{n}}^{(i) }$, which captures design concepts and complex referring relations from the generated explanations.
% Here, ${d_{\text{n}}}$ denotes the output dimensions of the text encoders.
Finally, the output $\bm{v}^{(i)}$ in VDFM is obtained as follows:
{\setlength{\abovedisplayskip}{4pt}
 \setlength{\belowdisplayskip}{4pt}
\begin{align}
\bm{v}^{(i)} = \text{TFLayers} \left( \left[ \bm{v}^{(i)}_{\text{s}}; \bm{v}^{(i)}_{\text{a}}; \bm{v}^{(i)}_{\text{n}} \right] \right).
\end{align}}%
% where ${d_{\text{img}}}$ denotes the output dimensions of VDFM.

\subsection{Confidence-Based Relaxed Alignment Module}
%4.10
% CRAM facilitates learning with unlabeled positives by estimating confidence scores and incorporates these scores into the loss function, which is motivated by the following facts.
Existing multimodal contrastive learning methods~\cite{Zhang2020ContrastiveLO, radford2021learning, Sun_2024_CVPR} primarily utilize contrastive loss functions, such as InfoNCE~\cite{oord2018representation}.
InfoNCE pairs each text with only one corresponding image as a positive sample, treating all other image-text pairs within the batch as negative samples.
% The model is then optimized to maximize similarity between corresponding positive samples and minimize similarity between negative samples.
% However, when the batch includes unlabeled positives, these are still treated as negatives.
As a result, certain pairs whose similarity should ideally be maximized are instead minimized, making single-label supervision inherently prone to noise~\cite{wu2022data}.
% While ReCo~\cite{Lin_2023_WACV} mitigates this issue by relaxing the constraints on negative pairs, ignoring those with similarity scores below zero, it does not explicitly handle unlabeled positives.
Therefore, we introduce a method for identifying unlabeled positives by estimating confidence scores for each sample.
We then incorporate these scores into a loss function that explicitly accounts for the unlabeled positives.

\begin{figure}[t]
    \centering
    \includegraphics[width=0.95\linewidth]{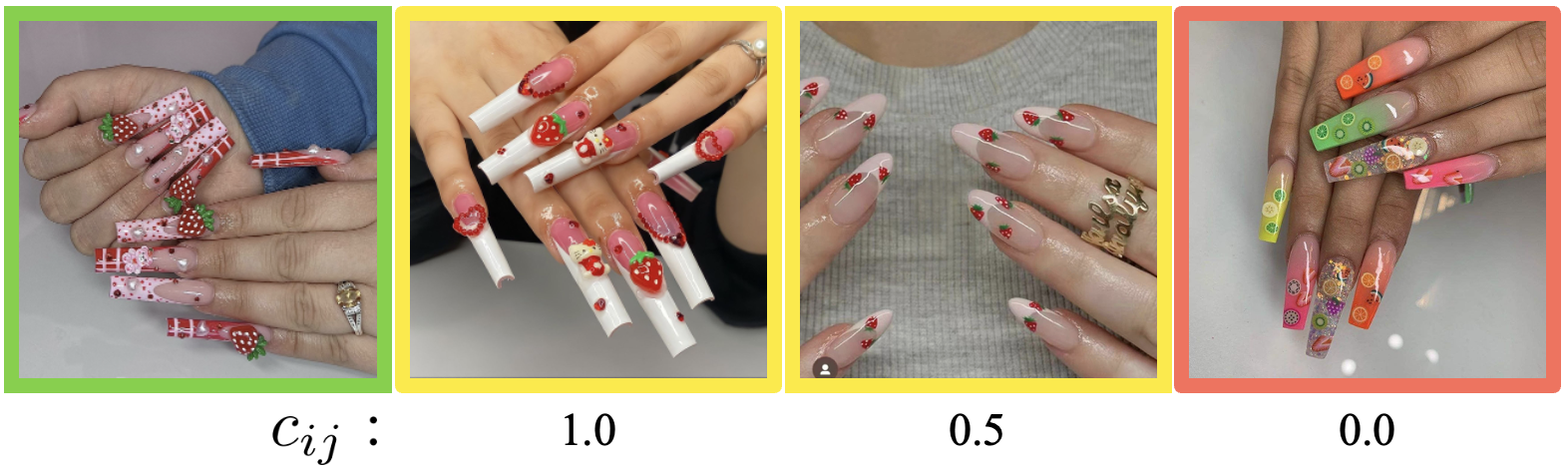}
    \vspace{-1mm}
    \caption{Examples of $c_{ij}$ estimated by the MLLM for the following $\bm{x}_{\text{txt}}$: ``Please put long fake nails on my nails and make them pink only at the base and the rest should be a fancy design with strawberries.'' Positives, unlabeled positives, and negatives are framed in green, yellow, and red, respectively.}
    \vspace{-4mm}
    \label{fig:confidence}
\end{figure}

We estimate the confidence scores $c_{ij} \in [0,1]$, which represent the degree to which $\bm{x}_{\text{img}}^{(j)}$ can be considered an unlabeled positive for $\bm{x}_{\text{txt}}^{(i)}$, leveraging an MLLM with a prompt $\bm{x}_\text{prompt}$, as follows:
{\setlength{\abovedisplayskip}{4pt}
 \setlength{\belowdisplayskip}{4pt}
\begin{align}
c_{ij} = \text{MLLM}( \bm{x}_\text{NNP}^{(i)}, \bm{x}_\text{img}^{(j)}, \bm{x}_\text{NNP}^{(j)}, \bm{x}_\text{prompt}),
\end{align}}%
\textcolor{blue}{where $\bm{x}_\text{NNP}^{(i)}$ denotes the normalized noun phrase corresponding to $\bm{x}_\text{txt}^{(i)}$.}
If $c_{ij} \geq \theta$, then $\bm{x}_{\text{img}}^{(j)}$ is considered to be an unlabeled positive for $\bm{x}_{\text{txt}}^{(i)}$, and $(i,j)$ is added to the set of unlabeled positive samples $\mathcal{Z}$, where $\theta$ is a threshold value.
\cref{fig:confidence} shows examples of $c_{ij}$.
Images that are closely aligned with $\bm{x}_\text{txt}$ are assigned higher scores, whereas those that only partially match $\bm{x}_\text{txt}$ receive lower scores.
The motivation for using $\bm{x}_\text{NNP}^{(j)}$ in addition to $\bm{x}_\text{NNP}^{(i)}$ and $\bm{x}_{\text{img}}^{(j)}$ is to make the MLLM focus more on the differences in design.
In fact, when using only $\bm{x}_\text{NNP}^{(i)}$ and $\bm{x}_{\text{img}}^{(j)}$  as inputs, both the description and the image focus on nails, which can lead to a situation in which even significantly different nail designs receive undesirably high confidence scores.
% By using $\bm{x}_\text{nnp}^{(j)}$ as a reference description, the differences between designs are explicitly represented in textual form, which helps prevent the assignment of excessively high scores to significantly different designs.
This method can be broadly applied to other multimodal retrieval tasks that contain a large number of similar images (e.g., \cite{Wu_2021_CVPR, 8099810}).
Appendix \ref{sec:confidence} describes a preprocessing method to speed up the above procedure.

%4.12
We next introduce the Confidence-based Relaxed Contrastive (CRC) loss for relaxing the contrastiveness using confidence scores.
The CRC loss is  defined as follows:
{\setlength{\abovedisplayskip}{4pt}
 \setlength{\belowdisplayskip}{4pt}
\begin{align}
\mathcal{L}_{\mathrm{CRC}} &= \mathcal{L}_{\mathrm{P}} + \lambda_{\mathrm{UP}} \mathcal{L}_{\mathrm{UP}} + \lambda_{\mathrm{N}} \mathcal{L}_{\mathrm{N}}, \\
\mathcal{L}_{\mathrm{P}} &= \sum\nolimits_{i} \left( 1 - S_{ii} \right)^2, \\
\mathcal{L}_{\mathrm{UP}} &= \sum\nolimits_{\substack{(i,j) \in \mathcal{Z}}} \left(\max(c_{ij} - S_{ij}, 0)\right)^2, \\
\mathcal{L}_{\mathrm{N}} &= \sum\nolimits_{\substack{(i,j) \notin \mathcal{Z}}} \left(\max(S_{ij}, 0 )\right)^2,
\end{align}}%
where $\lambda_{\mathrm{UP}}$ and $\lambda_{\mathrm{N}}$ are hyperparameters that control the weights of respective loss terms.
Terms $\mathcal{L}_{\mathrm{P}}$, $\mathcal{L}_{\mathrm{UP}}$, and $\mathcal{L}_{\mathrm{N}}$ denote the loss terms corresponding to positive, unlabeled positive, and negative samples, respectively.
% Similarity $S_{ij}$ is obtained by substituting $\bm{l}'^{(i)}$ with $\bm{l}^{(i)}$ and $\bm{v}'^{(j)}$ with $\bm{v}^{(j)}$ in \Cref{eq:sim}.
% Their similarities $S'_{ij}$ are computed as the cosine similarity~\cite{radford2021learning} between $\bm{l}'^{(i)}$ and $\bm{v}'^{(j)}$.
We calculate the cosine similarity $S_{ij}$ between $\bm{l}_\text{+}^{(i)}$ and $\bm{v}^{(j)}$.
On the basis of $\mathcal{Z}$, a penalty $(c_{ij} - S_{ij})^2$ is applied to samples where $(i, j) \in \mathcal{Z}$ and $S_{ij} < c_{ij}$.
Specifically, in the max function of $\mathcal{L}_{\mathrm{UP}}$, only samples with similarity scores below $c_{ij}$ contribute to the loss, whereas samples with similarity scores exceeding $c_{ij}$ are ignored. 
%4.11
The output of the model during inference is $\hat{Y}$, which is a list of nail design images, by sorting $X_{\text{img}}$ in descending order based on $S_{ij}$. 
\vspace{-1mm}
\section{NAIL-STAR Benchmark}
\vspace{-1mm}
%5.1
\textcolor{blue}{
We constructed a new benchmark for the NAIL-STAR task, because few datasets jointly include descriptions, palette queries, and images.
Indeed, existing datasets that include nail images~\cite{fan2024segmentation, nails-segmentation-vhnmw-dataset, segment-ca1h0-u1ri1-dataset} primarily focus on the segmentation task, making them unsuitable for the NAIL-STAR task.}
Moreover, the largest dataset in related fields contains fewer than 1,600 images of nails painted in a single color, which is insufficient.
% Furthermore, the study by Fan et al.~\cite{fan2024segmentation} explicitly excludes all nail design images.
% As a result, existing datasets provide only a limited number of images that are practically useful for nail design image retrieval. 
\textcolor{blue}{We collected 10,625 nail design images and annotated them with detailed, multi-layered descriptions and palette queries.
Details on image filtering, annotation, word frequency distribution of descriptions, and color distribution of nail designs are provided in Appendix~\ref{sec:nailster}.
}
The NAIL-STAR benchmark offers three main contributions:
\begin{itemize}
    \setlength{\parskip}{0.2mm} % 段落間
    \setlength{\itemsep}{0.2mm} % 項目間
    % \item Nail design often consists of a painted part, which allows for creative flexibility, and a decorative part, which can be modified only by selecting and arranging pre-manufactured embellishments. Therefore, the NAIL-STAR benchmark differs from datasets consisting of prompt and generated image pairs or datasets composed solely of pre-existing products (e.g., fashion~\cite{rostamzadeh2018fashion, Han_2017_ICCV, Wu_2021_CVPR}).

    \item The NAIL-STAR benchmark is unique as it focuses on dense intent descriptions, in contrast to existing datasets for multimodal retrieval (e.g., \cite{rostamzadeh2018fashion, Han_2017_ICCV, Wu_2021_CVPR, lin2014microsoft, Plummer_2015_ICCV}) that are typically based on simple, short phrases. 
    \item \textcolor{blue}{The NAIL-STAR benchmark includes palette queries consisting of zero or more colors specified as continuous values in the RGB space via a color picker. This contrasts with the text-labeled color attributes used in existing fashion datasets (e.g., \cite{rostamzadeh2018fashion, Liu_2016_CVPR}).}
    \item The images were curated from user posts across various countries and regions via Pinterest. Given the 42 languages used by posting users, the benchmark is expected to include images from users across diverse regions.
\end{itemize}

\vspace{-3mm}
\paragraph{Statistics.}
%5.5
% The NAIL-STAR benchmark consists of 10,625 nail design images and English descriptions.
\textcolor{blue}{The NAIL-STAR benchmark consists of 10,625 nail design images with English descriptions and palette queries.}
The descriptions were given by 208 annotators, with an average of 51.1 descriptions per annotator.
It has a vocabulary size of 7,014, a total of 228,080 words, and an average sentence length of 21.5 words.
\textcolor{blue}{The palette queries were automatically derived for the training and validation sets, and manually annotated for the test set.
The average number of colors per palette query is 2.0 in the training and validation sets and 2.1 in the test set.
}
%5.6
The training, validation, and test sets consist of 8,625, 400, and 1,600 samples, respectively. 
%5.7
We used the training set for training models, the validation set for adjusting their hyperparameters, and the test set for evaluating the models.
% The NAIL-STAR benchmark is detailed in Appendix \ref{sec:nailster}.

\begin{table*}[!ht]
    \centering
    \small
    \begin{tabular}{lccccccc}
        \toprule
        \multirow{2}{*}{Method} & \multirow{2}{*}{w/$\bm{x}_\text{pal}$} & \multicolumn{3}{c}{NAIL-STAR} & \multicolumn{3}{c}{Marqo Fashion200K}\\
        \cmidrule(lr){3-5} \cmidrule(lr){6-8}
        & & MRR↑ [\%] & R@1↑ [\%] & R@10↑ [\%] & MRR↑ [\%] & R@1↑ [\%] & R@10↑ [\%] \\
        \hline
        CLIP~\cite{radford2021learning} & N & 25.2 \scriptsize{$\pm$ 0.1} & 15.5 \scriptsize{$\pm$ 0.3} & 44.4 \scriptsize{$\pm$ 0.3} & 61.7 \scriptsize{$\pm$ 0.5} & 47.6 \scriptsize{$\pm$ 0.7} & 88.8 \scriptsize{$\pm$ 0.1} \\
        FashionViL~\cite{han2022fashionvil} & N & 20.8 \scriptsize{$\pm$ 0.6} & 11.1 \scriptsize{$\pm$ 0.9} & 40.9 \scriptsize{$\pm$ 1.4} & 61.7 \scriptsize{$\pm$ 0.5} & 47.3 \scriptsize{$\pm$ 0.5} & 89.1 \scriptsize{$\pm$ 0.5}\\
        FAME-ViL~\cite{Han_2023_CVPR} & N & 45.1 \scriptsize{$\pm$ 0.3} & 31.7 \scriptsize{$\pm$ 0.7} & 71.8 \scriptsize{$\pm$ 0.3} & 65.3 \scriptsize{$\pm$ 0.5} & 51.9 \scriptsize{$\pm$ 0.8} & 90.5 \scriptsize{$\pm$ 0.3}\\
        BEiT-3~\cite{Wang_2023_CVPR} & N & 53.9 \scriptsize{$\pm$ 0.4} & 40.6 \scriptsize{$\pm$ 0.6} & 79.8 \scriptsize{$\pm$ 0.2} & 66.2 \scriptsize{$\pm$ 0.8} & 52.8 \scriptsize{$\pm$ 0.9} & 90.9 \scriptsize{$\pm$ 0.5}\\
        BLIP-2~\cite{pmlr-v202-li23q} & N & 33.3 \scriptsize{$\pm$ 1.0} & 20.8 \scriptsize{$\pm$ 0.9} & 58.7 \scriptsize{$\pm$ 1.2} & 75.3 \scriptsize{$\pm$ 0.1} & 65.2 \scriptsize{$\pm$ 0.1} & 94.9 \scriptsize{$\pm$ 0.1}\\
        SigLIP~\cite{10377550} & N & 58.8 \scriptsize{$\pm$ 0.1} & 47.5 \scriptsize{$\pm$ 0.1} & 82.1 \scriptsize{$\pm$ 0.2} & 71.9 \scriptsize{$\pm$ 0.2} & 60.3 \scriptsize{$\pm$ 0.2} & 93.1 \scriptsize{$\pm$ 0.1}\\
        Alpha-CLIP~\cite{Sun_2024_CVPR}\footnotemark[2] & N & 27.1 \scriptsize{$\pm$ 0.6} & 16.1 \scriptsize{$\pm$ 0.5} & 50.3 \scriptsize{$\pm$ 1.0} & 59.8 \scriptsize{$\pm$ 0.3} & 45.5 \scriptsize{$\pm$ 0.4} & 87.4 \scriptsize{$\pm$ 0.2}\\
        Long-CLIP~\cite{zhang2025long} & N & 23.3 \scriptsize{$\pm$ 0.2} & 13.0 \scriptsize{$\pm$ 0.3} & 45.0 \scriptsize{$\pm$ 0.3} & 63.0 \scriptsize{$\pm$ 0.3} & 49.1 \scriptsize{$\pm$ 0.3} & 89.2 \scriptsize{$\pm$ 0.2}\\
        LamRA~\cite{liu2025lamra}\footnotemark[3] & N & 43.9 \scriptsize{$\pm$ 0.2} & 35.1 \scriptsize{$\pm$ 0.3} & 61.1 \scriptsize{$\pm$ 0.4} & 33.3 \scriptsize{$\pm$ 0.3} & 22.6 \scriptsize{$\pm$ 0.2} & 57.1 \scriptsize{$\pm$ 0.4}\\
        \rowcolor{pink!20}
        NaiLIA (description-only) & N & \underline{61.0} \scriptsize{$\pm$ 0.3} & \underline{49.5} \scriptsize{$\pm$ 0.4} & \underline{83.4} \scriptsize{$\pm$ 0.2} &
        \underline{82.0} \scriptsize{$\pm$ 0.2} & \underline{73.8} \scriptsize{$\pm$ 0.3} & \underline{96.3} \scriptsize{$\pm$ 0.1} \\
        \rowcolor{pink!20}
        NaiLIA (full) & Y & \textbf{67.6} \scriptsize{$\pm$ 0.5} & \textbf{56.4} \scriptsize{$\pm$ 0.7} & \textbf{88.0} \scriptsize{$\pm$ 0.3} & \textbf{82.7} \scriptsize{$\pm$ 0.3} & \textbf{74.6} \scriptsize{$\pm$ 0.2} & \textbf{96.9} \scriptsize{$\pm$ 0.2} \\
        \bottomrule
    \end{tabular}
    \vspace{-2mm}
    \caption{\textcolor{blue}{Quantitative comparison of NaiLIA and baseline methods on the NAIL-STAR and Marqo Fashion200K~\cite{marqo_fashion200k} benchmarks. All models were fine-tuned on their respective training sets. Here, w/$\bm{x}_\text{pal}$ indicates whether the model takes palette queries as input. The scores present the average and standard deviation of five trials. The highest score for each metric is in \textbf{bold}, and the second-highest score is \underline{underlined}. The results in the zero-shot setting are presented in Appendix \ref{sec:zero-shot}.}}
    \vspace{-4.5mm}
    \label{tab:quantitative}
\end{table*}

\section{Experiments}
\subsection{Experimental Setup}
\paragraph{Model Architecture.}
%5.13
In IPFM, we employed GPT-4o as the LLM and utilized BEiT-3~\cite{Wang_2023_CVPR} and SigLIP~\cite{10377550} as the text encoders.
In VDFM, we used DINOv2~\cite{Jose2024DINOv2MT} as the unimodal image encoder, and BEiT-3 and SigLIP as multimodal image encoders.
Additionally, to obtain the img2txt intent structure representation, we utilized GPT-4o and Qwen2-VL~\cite{Qwen2VL} as the MLLMs, and BEiT-3 and SigLIP as text encoders.
In CRAM, we employed Qwen2-VL as the MLLM to estimate the confidence scores.
\textcolor{blue}{The implementation details, including the training configuration, optimizer settings, computational environment, and feature dimensions, are provided in Appendix~\ref{sec:implementation}.}

\vspace{-4mm}
\paragraph{Benchmarks.}
\textcolor{blue}{
We conducted experiments on the NAIL-STAR benchmark and Marqo Fashion200K~\cite{marqo_fashion200k}, an extended version of the Fashion200K dataset~\cite{Han_2017_ICCV}.
% To the best of our knowledge, Marqo Fashion200K is the only fashion AI dataset designed for retrieval based on fine-grained, long-form textual queries, and we used this dataset. 
We used the Marqo Fashion200K because it is a fashion AI dataset designed for retrieval based on fine-grained, long-form textual queries.
Following the experimental protocol of Wang et al.~\cite{wang2024advancingmyopiaholismfully}, we used 5,000 samples for the test set, and 191,559 and 5,000 samples for the training and validation sets, respectively.
Since this dataset does not include palette queries, we added them; the details are in Appendix \ref{sec:palette-annotation}.
}

\vspace{-4mm}
\paragraph{Baselines.}
%6.2
% We used CLIP (ViT-B/32)~\cite{radford2021learning},  BLIP-2 (ViT-g)~\cite{pmlr-v202-li23q}, BEiT-3 (large)~\cite{Wang_2023_CVPR}, Alpha-CLIP (ViT-L/14)~\cite{Sun_2024_CVPR}, and Long-CLIP (ViT-L/14)~\cite{zhang2025long}, as the baseline methods.
% Furthermore, we fine-tuned both the text encoder and image encoder of CLIP (ViT-B/32) as an additional baseline.
We used CLIP (ViT-L/14)~\cite{radford2021learning},  FashionViL~\cite{han2022fashionvil}, FAME-ViL~\cite{Han_2023_CVPR}, BEiT-3 (large), BLIP-2 (ViT-g)~\cite{pmlr-v202-li23q}, SigLIP (L/16), Alpha-CLIP (ViT-L/14)~\cite{Sun_2024_CVPR}, Long-CLIP (ViT-L/14)~\cite{zhang2025long}, and LamRA~\cite{liu2025lamra} as the baseline methods.
%6.3
% CLIP, BLIP-2, BEiT-3, and Long-CLIP were selected because they are representative methods for zero-shot text-to-image retrieval tasks.
% In addition, Alpha-CLIP was selected for its use of masks, which enable multimodal retrieval with a focus on nails~\cite{Sun_2024_CVPR}.
% We also provided nail masks generated from $X_{\text{img}}$ using the segmentation model for fingernails~\cite{nakhoon_dataset} alongside $\bm{x}$ for Alpha-CLIP.
These models were selected as representative vision--language foundation models, including those tailored for the fashion domain.
\textcolor{blue}{All baseline methods were fine-tuned and took only $\bm{x}_{\text{txt}}$ and $X_{\text{img}}$ as inputs, without incorporating $\bm{x}_{\text{pal}}$.
The discussion on token length limitations of the proposed and baseline methods is provided in Appendix \ref{sec:token_length_limitation}.
}

\subsection{Quantitative Comparison}
%6.1
\Cref{tab:quantitative} shows the quantitative results for the performance of the proposed and baseline methods \textcolor{blue}{on the NAIL-STAR and Marqo Fashion200K benchmarks.
For a fair comparison with the baselines, we report results for NaiLIA with and without $\bm{x_{\text{pal}}}$ as input.}
% We conducted experiments with CLIP (freezed)~\cite{radford2021learning}, BLIP-2~\cite{pmlr-v202-li23q}, BEiT-3~\cite{Wang_2023_CVPR}, Alpha-CLIP~\cite{Sun_2024_CVPR}, and Long-CLIP~\cite{zhang2025long} in a zero-shot setting.
% The scores for these models were based on a single trial because these provide consistent results across multiple trials.
The scores present the average and standard deviation from five trials.
We used mean reciprocal rank (MRR) and recall@$K$ ($K=1,10$) as the evaluation metrics, with recall@1 as the primary metric because these are standard metrics for image retrieval~\cite{han2022fashionvil, Song_2024_CVPR}.
% \textcolor{blue}{The highest score for each metric is in \textbf{bold}, and the second-highest score is \underline{underlined}.}

%6.6
% \Cref{tab:quantitative} shows that the proposed method achieved the highest recall@1 score of 49.5$\%$, outperforming the baselines (i), (ii), (iii), (iv), (v), (vi), (vii), (viii), and (ix) with improvements of 34.0, 38.4, 17.8, 8.9, 28.7, 2.0,  33.4, 36.5, and 39.4 points, respectively.
\textcolor{blue}{
On the NAIL-STAR benchmark, \Cref{tab:quantitative} shows that the proposed full model achieved the highest recall@1 score of 56.4\%, outperforming the best baseline, SigLIP, by 8.9 points.
The description-only model also surpassed this baseline by 2.0 points, suggesting that NaiLIA maintains robust retrieval performance even when users do not provide $\bm{x}_\text{pal}$.
On the Marqo Fashion200K, our full model achieved the highest recall@1 score of 74.6\%, exceeding the best baseline, BLIP-2, by 9.4 points.
The description-only variant also demonstrated an improvement of 8.6 points over this baseline.}
Moreover, NaiLIA outperformed the baseline methods across other metrics.
%6.7
The performance differences between our method and baseline methods were statistically significant ($p$-value $< 0.01$).

\footnotetext[2]{We provided segmentation masks from $X_{\text{img}}$ using a fingernail segmentation model~\cite{nakhoon_dataset} for the NAIL-STAR benchmark and a segmentation model~\cite{Xie2021SegFormerSA} fine-tuned on a fashion dataset for the Marqo Fashion200K.}
\footnotetext[3]{Scores are reported after fine-tuning LamRA-Ret and LamRA-Rank.}

\begin{figure*}[t]
    \centering
    \includegraphics[width=\linewidth]{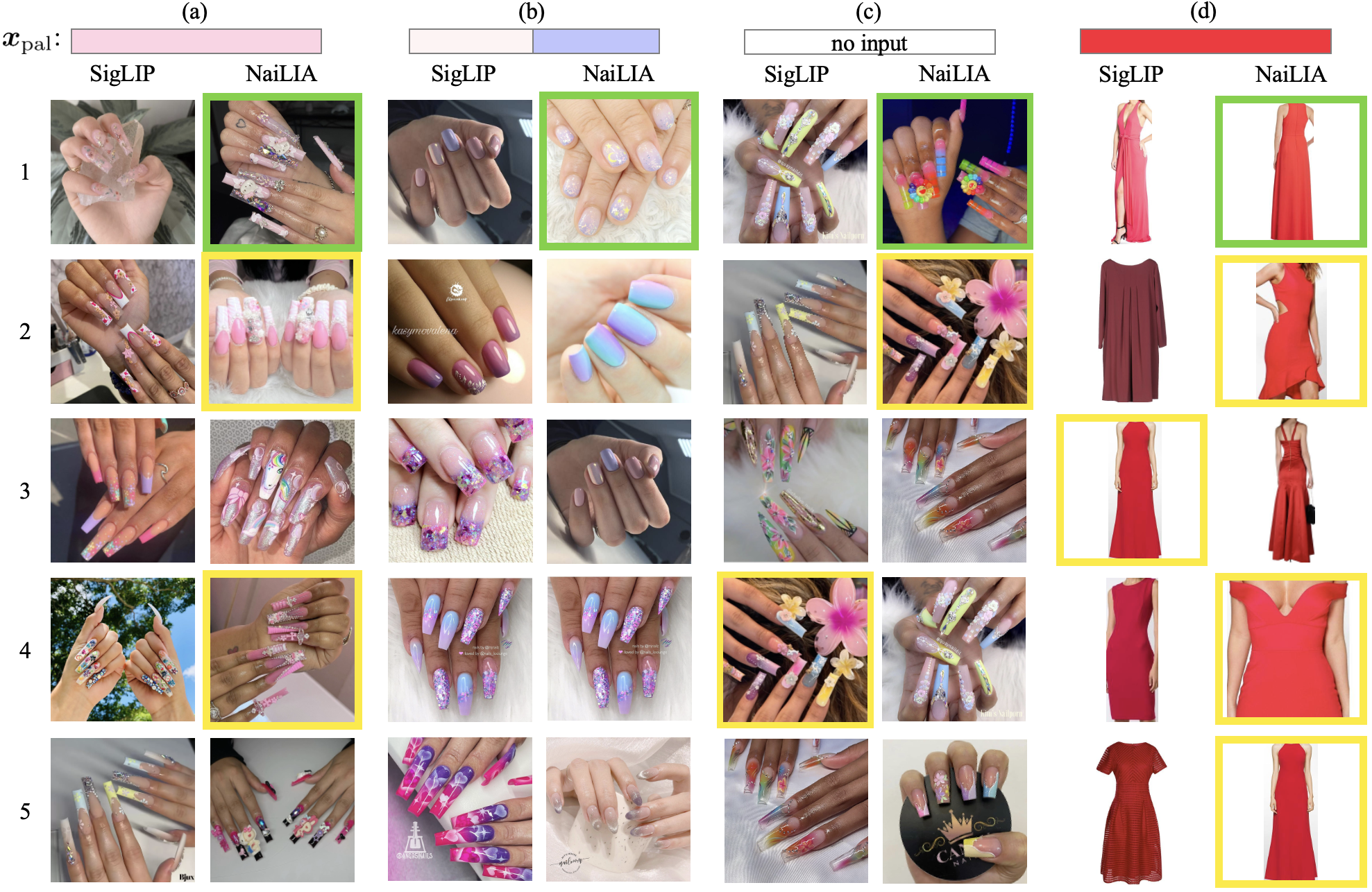}
    \caption{\textcolor{blue}{Qualitative results of the proposed method (NaiLIA) and a baseline method (SigLIP)~\cite{10377550} for the following $\bm{x}_{\text{txt}}$ and $\bm{x}_{\text{pal}}$: (a) ``I'd like my nails to have a cute, teenage vibe. I'd love a pink base with floral patterns and maybe some character accessories. Can we do a long nail shape?'' with \colorpatchframe{255}{211}{229} (\#ffd3e5). (b) ``I would like a fairy-tale nail design based on purple and pink.'' with \colorpatchframe{255}{244}{244} (\#fff4f4) and \colorpatchframe{191}{197}{255} (\#bfc5ff). (c) ``I'd like a colorful and flashy nail design. Please add a large flower nail stone to the ring finger. The tips of the nail tips should be square-shaped.'' in the setting without $\bm{x}_{\text{pal}}$. (d) ``red dress with a high waist. the dress is made of a stretchy material and has a flowy skirt. the dress is a formal style and is suitable for a special occasion.'' with \colorpatchframe{255}{31}{54} (\#ff1f35). The top-5 retrieved images are shown. Positives and unlabeled positives are enclosed in green and yellow frames, respectively. Examples (a), (b), and (c) show successful cases of NaiLIA on the NAIL-STAR benchmark, whereas (d) is a successful case on the Marqo Fashion200K benchmark. Additional qualitative results, including cases where $\bm{x}_{\text{pal}}$ consists of more than three colors, are presented in Appendix~\ref{sec:additonal_qualitative}.} }
    \label{fig:qualitative}
    \vspace{-4mm}
\end{figure*}

\subsection{Qualitative Results}
\vspace{-1mm}
\label{sec:qualitative}
%6.8
\cref{fig:qualitative} shows the qualitative results of NaiLIA and the SigLIP~\cite{10377550} baseline method. 
The top-5 retrieved images are shown.
Positives and unlabeled positives are in green and yellow frames, respectively.
\textcolor{blue}{\cref{fig:qualitative} (a), (b), and (c) show successful cases of NaiLIA on the NAIL-STAR benchmark, while (d) presents one on the Marqo Fashion200K.}

\textcolor{blue}{\cref{fig:qualitative} (a) shows an example for the following $\bm{x}_{\text{txt}}$ and $\bm{x}_{\text{pal}}$: \textit{``I'd like my nails to have a cute, teenage vibe. I'd love a pink base with floral patterns and maybe some character accessories. Can we do a long nail shape?''} with \colorpatchframe{255}{211}{229} (\#ffd3e5).}
NaiLIA successfully retrieved the target nail design image as the top result. 
Although the 2nd and 4th retrieved images did not fully incorporate ``floral patterns,'' they were still largely aligned with the description as unlabeled positives. 
\textcolor{blue}{This demonstrates the model's ability to comprehend abstract expressions related to impressions such as ``cute, teenage vibe'' in addition to character decorations and subtle tonal nuances of pink.}
In contrast, the baseline method retrieved the target image at rank 24.

\textcolor{blue}{
Another instance where NaiLIA performed well is shown in \cref{fig:qualitative} (b).
In this case, $\bm{x}_{\text{txt}}$ and $\bm{x}_{\text{pal}}$ were \textit{``I would like a fairy-tale nail design based on purple and pink.''} with \colorpatchframe{255}{244}{244} (\#fff4f4) and \colorpatchframe{191}{197}{255} (\#bfc5ff).
NaiLIA retrieved the target nail design image at rank 1 by comprehending that the star and moon decorations, along with the gradient finish, evoke ``a fairy-tale'' impression.
Furthermore, it ranked nail designs with pastel purple and/or pink tones that closely align with $\bm{x}_{\text{pal}}$ within the top-5 results.
In contrast, although the baseline method retrieved designs with purple and/or pink tones at high ranks, some had lower brightness than specified in $\bm{x}_{\text{pal}}$, ranking the target nail design at 31.
}

\textcolor{blue}{\cref{fig:qualitative} (c) presents the results in the description-only setting, where $\bm{x}_{\text{pal}}$ is not provided.}
Here, $\bm{x}_{\text{txt}}$ was \textit{``I'd like a colorful and flashy nail design. Please add a large flower nail stone to the ring finger. The tips of the nail tips should be square-shaped.''}
Despite the apparent visual dissimilarity between the flower-shaped decoration in the target nail design image and a real flower, NaiLIA successfully recognized that this decoration symbolizes a flower and retrieved the target nail design image as the top result.
Note that ``a large flower nail stone'' in $\bm{x}_{\text{txt}}$ does not necessarily mean that the user intends the decoration to closely resemble a real flower. 
The user instead may be referring to a flower-inspired character decoration, as shown in the target nail design image.
% Furthermore, the unlabeled positive that meets the conditions related to nail accessories, nail shape, and design impressions such as ``colorful and flash'' was retrieved at rank 2, although it does not conform to the specification for the ring finger.
By contrast, the baseline method retrieved the target nail design image at rank 137.

%5.9
\textcolor{blue}{\cref{fig:qualitative} (d) shows an example on the Marqo Fashion200K.
Here, $\bm{x}_{\text{txt}}$ and $\bm{x}_{\text{pal}}$ were \textit{``red dress with a high waist. the dress is made of a stretchy material and has a flowy skirt. the dress is a formal style and is suitable for a special occasion.''} with \colorpatchframe{255}{31}{54} (\#ff1f35). 
In this example, NaiLIA retrieved the target image at rank 1 and also retrieved unlabeled positives aligned with both $\bm{x}_{\text{txt}}$ and $\bm{x}_{\text{pal}}$ at ranks 2, 4, and 5.
In contrast, although the top-5 results from the baseline were generally consistent with $\bm{x}_{\text{txt}}$, most of them exhibited red tones that were either more pinkish or deeper than the color specified in $\bm{x}_{\text{pal}}$.
As a result, the baseline ranked the target image at 37.}

\begin{table*}[!ht]
    \centering
    \small
    \begin{tabular}{cccccccccccc}
        \toprule
        \multirow{2}{*}{Model} & \multicolumn{2}{c}{IPFM} & \multicolumn{3}{c}{VDFM} & \multicolumn{1}{c}{CRAM} & \multirow{2}{*}{MRR↑ [\%]} & \multirow{2}{*}{R@1↑ [\%]} & \multirow{2}{*}{R@10↑ [\%]} \\
        \cmidrule(lr){2-3} \cmidrule(lr){4-6} \cmidrule(lr){7-7} 
         & $\bm{l}_{\text{MDD}}$ & $\bm{l}_{\text{NNP}}$ & $\bm{v}_\text{s}$ & $\bm{v}_\text{a}$ & $\bm{v}_\text{n}$ & loss & & & & \\
        \hline
        (a) & \cmark & \cmark & \cmark & \cmark & \cmark & CRC & \textbf{67.6} \scriptsize{$\pm$ 0.5} & \textbf{56.4} \scriptsize{$\pm$ 0.7} & \textbf{88.0} \scriptsize{$\pm$ 0.3} \\
        \hline
        (b) &  & \cmark & \cmark & \cmark & \cmark & CRC & 66.0 \scriptsize{$\pm$ 0.5} & 54.9 \scriptsize{$\pm$ 0.7} & 87.1 \scriptsize{$\pm$ 0.2} \\
        (c) & \cmark &  & \cmark & \cmark & \cmark & CRC & 65.6 \scriptsize{$\pm$ 0.8} & 54.5 \scriptsize{$\pm$ 0.9} & 87.2 \scriptsize{$\pm$ 0.6} \\
        (d) &  &  & \cmark & \cmark & \cmark & CRC & 63.0 \scriptsize{$\pm$ 0.5} & 51.6 \scriptsize{$\pm$ 0.6} & 84.8 \scriptsize{$\pm$ 0.5} \\
        \hline
        (e) & \cmark & \cmark &  & \cmark & \cmark & CRC & 66.4 \scriptsize{$\pm$ 0.4} & 55.1 \scriptsize{$\pm$ 0.7} & 87.3 \scriptsize{$\pm$ 0.3} \\
        (f) & \cmark & \cmark & \cmark &  & \cmark & CRC & 53.9 \scriptsize{$\pm$ 0.2} & 42.1 \scriptsize{$\pm$ 0.1} & 77.4 \scriptsize{$\pm$ 0.3} \\
        (g) & \cmark & \cmark & \cmark & \cmark &  & CRC & 65.4 \scriptsize{$\pm$ 0.4} & 54.0 \scriptsize{$\pm$ 0.6} & 87.2 \scriptsize{$\pm$ 0.6} \\
        (h) & \cmark & \cmark & & \cmark &  & CRC & 64.6 \scriptsize{$\pm$ 0.5} & 53.1 \scriptsize{$\pm$ 0.6} & 86.8 \scriptsize{$\pm$ 0.3} \\
        \hline
        (i) & \cmark & \cmark & \cmark & \cmark & \cmark & InfoNCE & 64.3 \scriptsize{$\pm$ 0.4} & 52.7 \scriptsize{$\pm$ 0.7} & 86.4 \scriptsize{$\pm$ 0.4} \\
        (j) & \cmark & \cmark & \cmark & \cmark & \cmark & CRC ($\lambda_{\mathrm{UP}}=0$) & 65.7 \scriptsize{$\pm$ 0.6} & 54.5 \scriptsize{$\pm$ 0.6} & 87.0 \scriptsize{$\pm$ 0.3} \\
        (k) & \cmark & \cmark & \cmark & \cmark & \cmark & CRC ($c_{ij} = 0.7$) & 66.0 \scriptsize{$\pm$ 0.8} & 55.1 \scriptsize{$\pm$ 0.9} & 87.1 \scriptsize{$\pm$ 0.2} \\
        \bottomrule
    \end{tabular}
    \vspace{-1mm}
    \caption{Ablation study on the core modules in our method. The average and standard deviation from five trials are reported. The highest score for each metric is in \textbf{bold}. The additional ablation study is presented in Appendix \ref{sec:vdfm_ablation}.}
    \vspace{-3mm}
    \label{tab:ablation}
\end{table*}

\begin{table}[!ht]
    \centering
    \small
    \setlength{\tabcolsep}{3pt}
    \resizebox{\columnwidth}{!}{
    \begin{tabular}{lccccc}
        \toprule
        Model & MRR↑ [\%] & R@1↑ [\%] & R@10↑ [\%] \\
        \hline
        CLIP~\cite{radford2021learning} w/ InfoNCE & 25.2 \scriptsize{$\pm$ 0.1} & 15.5 \scriptsize{$\pm$ 0.3} & 44.4 \scriptsize{$\pm$ 0.3} \\
        CLIP~\cite{radford2021learning} w/ CRC loss & 25.9 \scriptsize{$\pm$ 0.1} & 16.5 \scriptsize{$\pm$ 0.3} & 44.8 \scriptsize{$\pm$ 0.6} \\
        \bottomrule
    \end{tabular}
    }
    \vspace{-1mm}
    \caption{\textcolor{blue}{Quantitative comparison of CLIP fine-tuned with InfoNCE vs. CRC loss on the NAIL-STAR benchmark.}}
    \vspace{-3mm}
    \label{tab:clip_crc}
\end{table}

\subsection{Ablation Study}
\vspace{-1mm}
%5.10
\Cref{tab:ablation} shows the results of ablation studies. 
The table presents the average and standard deviation of the results of five trials.
% The highest score for each metric is in bold.
% The ablation conditions were as follows:

\vspace{-3mm}
\paragraph{IPFM ablation.}
While keeping the Transformer layers, we removed $\bm{l}_{\text{MDD}}$ and/or $\bm{l}_{\text{NNP}}$ from IPFM to assess their contributions.  
% The results show decreases of 1.5 points and 2.1 points in recall@10 for Models (b) and (c), respectively, compared with the findings for Model (a).
The results show decreases of 1.5, 1.9, and 4.8 points in recall@1 for Models (b), (c), and (d), respectively, compared with the findings for Model (a).
This indicates the effectiveness of representing dense intent descriptions structurally by using normalized noun phrases and multi-layered design descriptions in addition to the original descriptions.

\vspace{-4mm}
\paragraph{VDFM ablation.}
% Similarly, we removed  $\bm{v}_\text{s}$, $\bm{v}_\text{a}$, or $\bm{v}_\text{n}$ from the VDFM to investigate their respective effects on performance. 
Similarly, we removed either single components or pairs of  $\bm{v}_\text{s}$, $\bm{v}_\text{a}$, and $\bm{v}_\text{n}$ from VDFM to investigate their respective and combined effects on performance. 
% The results show decreases of 0.2 points, 5.1 points, and 3.8 points in recall@10 for Models (d), (e), and (f), respectively, compared with the findings for Model (a).
The results show decreases of 1.3, 14.3, 2.4, and 3.3 points in recall@1 for Models (e), (f), (g), and (h), respectively, compared with the findings for Model (a).
This suggests the img2txt intent structure representation contributed to modeling the multi-layered design concepts in $X_\text{img}$.

\vspace{-3mm}
\paragraph{CRC loss ablation.}
We changed the loss function to InfoNCE~\cite{oord2018representation} to investigate the effectiveness of the CRC loss.
Moreover, we ablated the $L_\text{UP}$ component of the CRC loss by setting $\lambda_{\mathrm{UP}} = 0$ to assess the contribution of this term, which constitutes the main novelty of the loss.
Furthermore, to evaluate the validity of score estimation by the MLLMs, we conducted an experiment in which we fixed $c_{ij} = 0.7$ instead of using the estimated values.
% The results show decreases of 1.4 points and 0.3 points in recall@10 for Models (g) and (h), respectively, compared with the findings for Model (a).
The results show decreases of 3.7, 1.9, and 1.3 points in recall@1 for Models (i), (j), and (k), respectively, compared with the findings for Model (a).
The results suggest that introducing a loss function to handle unlabeled positives and estimating confidence scores leads to an efficient training process.
% These results suggest that when a batch contains images that can be considered positive samples, handling them as unlabeled positives rather than as negative samples enables contrastive learning that is robust to similar images.

% Model (i) showed the largest performance drop compared to Model (a), apart from Model (f), for which performance degradation was expected, and Models (d) and (h), in which multiple types of features were ablated. 
% This suggests that replacing InfoNCE with the CRC loss contributed most significantly to the performance improvement.

\textcolor{blue}{Moreover, to demonstrate that the CRC loss is not exclusive to the NaiLIA architecture, we evaluated its effectiveness on a different model.
\Cref{tab:clip_crc} presents a quantitative comparison on the NAIL-STAR benchmark between two variants of the CLIP model: one fine-tuned with the conventional InfoNCE loss and the other with the proposed CRC loss.
The table presents the average and standard deviation of the results of five trials.
The results show that the model trained with the CRC loss achieves a 1.0 point improvement in recall@1 compared to the InfoNCE variant.
This implies that the CRC loss is not specific to NaiLIA and can serve as a general-purpose loss function for retrieval tasks.}

\begin{table}[!t]
    \centering
    \small
    \setlength{\tabcolsep}{4pt}
    \resizebox{\columnwidth}{!}{
    \begin{tabular}{lcccc}
        \toprule
        Method & w/$\bm{x}_\text{pal}$ & MRR↑ [\%] & R@1↑ [\%] & R@10↑ [\%] \\
        \hline
        CLIP~\cite{radford2021learning} & N & 25.2 \scriptsize{$\pm$ 0.1} & 15.5 \scriptsize{$\pm$ 0.3} & 44.4 \scriptsize{$\pm$ 0.3}\\
        CLIP~\cite{radford2021learning} & Y & 29.7 \scriptsize{$\pm$ 0.2} & 21.3 \scriptsize{$\pm$ 0.3} & 48.7 \scriptsize{$\pm$ 0.2} \\
        SigLIP~\cite{10377550} & N & 58.8 \scriptsize{$\pm$ 0.1} & 47.5 \scriptsize{$\pm$ 0.1} & 82.1 \scriptsize{$\pm$ 0.2} \\
        SigLIP~\cite{10377550} & Y & 63.3 \scriptsize{$\pm$ 0.1} & 53.4 \scriptsize{$\pm$ 0.1} & 86.2 \scriptsize{$\pm$ 0.1} \\
        \bottomrule
    \end{tabular}
    }
    \vspace{-1mm}
    \caption{\textcolor{blue}{Quantitative results of applying palette queries to CLIP and SigLIP on the NAIL-STAR benchmark.}}
    \label{tab:ablation_color}
\end{table}

\vspace{-3mm}
\paragraph{Palette ablation.}
\textcolor{blue}{
We conducted experiments on CLIP and SigLIP by providing $\bm{x}_\text{pal}$ as an input in addition to $\bm{x}_{\text{txt}}$ and $X_{\text{img}}$ on the NAIL-STAR benchmark.
As shown in \Cref{tab:ablation_color}, the results indicate that incorporating $\bm{x}_\text{pal}$ improved CLIP and SigLIP's performance by 5.8 and 5.9 points, respectively, compared with the setting without $\bm{x}_\text{pal}$.
This suggests that a palette query processing approach is broadly applicable to other vision--language foundation models and that incorporating it enhances retrieval performance.
}

\vspace{-1mm}
\section{Conclusion}
\vspace{-1mm}
%7.1
In this study, we focused on the NAIL-STAR task, which aims to retrieve nail design images that align with dense intent descriptions \textcolor{blue}{and palette queries}.
%7.2
% To build our multimodal retrieval model, NaiLIA, we introduced three modules: CRAM, VDFM, and LIFM.
NaiLIA estimates confidence scores of unlabeled positives and incorporates these scores into the loss function to enable model training with unlabeled positives. 
\textcolor{blue}{To capture multi-layered user intent and continuous multi-color preferences, it models interactions between dense intent descriptions and palette queries.}
To handle nail design images, it integrates (i) a unimodal visual representation, (ii) a multimodal aligned representation, and (iii) an img2txt intent structure representation.
% LIFM models dense intent descriptions by structuring them around dense intent to capture the multi-layered user intent embedded within.
% LIFM captures the multi-layered user intent in dense intent descriptions by modeling the descriptions structured in terms of dense intent.
% It is capable of modeling dense intent descriptions by capturing the multi-layered nature of intent descriptions through structured representations.
We constructed the NAIL-STAR benchmark, which consists of pairs of a dense intent description and a palette query, and their corresponding diverse nail design images.
Experimental results demonstrate that the proposed method outperforms standard methods.

{
    \small
    \bibliographystyle{ieeenat_fullname}
    \bibliography{main}
}

\clearpage
\setcounter{page}{1}
\maketitlesupplementary

\renewcommand{\thesection}{\Alph{section}}
\setcounter{section}{0}

In this supplementary material, we provide the prompts and generated texts used in each module of NaiLIA, along with details of the NAIL-STAR benchmark construction and data distributions. We also present additional quantitative and qualitative results, including ablation studies and error analyses. Furthermore, the supplementary material includes a project page, which readers may refer to for an overview of NaiLIA or to view qualitative results at a higher resolution.

\section{Additional Related Work}
\paragraph{Fashion AI.}
%2.2
In the domain of fashion AI, numerous studies have been conducted on various tasks, including clothing category recognition~\cite{Wang_2018_CVPR}, fashion image retrieval~\cite{Park_2019_CVPR_Workshops, Kuang_2019_ICCV}, virtual try-on~\cite{Han_2018_CVPR, choi2021viton, Ning_2024_CVPR}, fashion recommendation~\cite{Yu_2019_ICCV}, trend forecasting~\cite{Al-Halah_2017_ICCV}, and fashion compatibility~\cite{Pang_2024_WACV}.
Representative subfields in fashion AI include those related to makeup and fingernails~\cite{li2018beautygan, Jiang_2020_CVPR, Hu_2022_CVPR, fan2024segmentation, 10710729}.
In the field of makeup, makeup transfer~\cite{li2018beautygan, Jiang_2020_CVPR, Hu_2022_CVPR} aims to transfer the makeup style of a reference image to the source image.
In nail-related research, NailNet~\cite{fan2024segmentation} is a segmentation method that segments the fingernail and the lunula, which can be applied to virtual nail art.
Moreover, Yarimbiyik et al.~\cite{10710729} propose a model that recommends the nail shapes that suit the shape of the fingers.
Several studies have proposed fashion-centric vision-language foundation models~\cite{Zhuge_2021_CVPR, Han_2023_CVPR, Han_2023_CVPR_2, Song_2024_CVPR} that tackle downstream multimodal tasks, such as cross-modal retrieval~\cite{rostamzadeh2018fashion, FashionBERT}, category recognition~\cite{rostamzadeh2018fashion, Wang_2018_CVPR}, fashion image captioning~\cite{rostamzadeh2018fashion, XuewenECCV20Fashion}, and text-guided image retrieval~\cite{Wu_2021_CVPR}.

\paragraph{Contrastive learning.}
Contrastive learning is widely adopted in vision-language foundation models (e.g., \cite{radford2021learning, li2022blip}). 
CLIP employs InfoNCE~\cite{oord2018representation} as its contrastive loss function. 
However, training with InfoNCE assumes a strict one-to-one correspondence between a single text and a single image as a positive example. 
Consequently, samples that are semantically aligned but not explicitly labeled as positive pairs are treated as negatives, introducing noise into the learning process~\cite{wu2022data}.
To address this, various loss functions have been proposed to relax the one-to-one labeling constraint in InfoNCE-based training (e.g., \cite{wu2022data, Lin_2023_WACV, relaxformer, gao2022pyramidclip}).
ReCo~\cite{Lin_2023_WACV} mitigates this issue by relaxing the constraints on negative pairs, ignoring those with similarity scores below zero, but it does not explicitly handle unlabeled positives.

\paragraph{Benchmarks on fashion AI.}
As mentioned above, there are several benchmarks for multimodal retrieval in fashion AI~\cite{rostamzadeh2018fashion, Han_2017_ICCV, Wu_2021_CVPR}.
In the field of image synthesis, research has been conducted on tasks such as virtual try-on~\cite{Han_2018_CVPR, choi2021viton, Ning_2024_CVPR} and makeup transfer~\cite{li2018beautygan, Jiang_2020_CVPR, Hu_2022_CVPR}.
For instance, VITON-HD~\cite{choi2021viton} serves as a benchmark for evaluating high-resolution virtual try-on methods, focusing on garment detail preservation, misalignment handling, and realistic synthesis.
In addition, the Makeup Transfer dataset~\cite{li2018beautygan} is utilized to evaluate the effectiveness and robustness of makeup transfer techniques across various makeup styles and lighting conditions.

Several benchmarks focus on nail-related research~\cite{fan2024segmentation, nails-segmentation-vhnmw-dataset, segment-ca1h0-u1ri1-dataset, Duke2019NailPT}.
These studies primarily focus on segmentation tasks and are applied in various domains, including assisting disease assessment in medical diagnostics, forensic personal identification, and the beauty industry for virtual nail art applications~\cite{fan2024segmentation}.

\paragraph{Comparison with existing benchmarks.}
The NAIL-STAR benchmark constructed in this study differs from existing benchmarks in the following aspects.
First, existing benchmarks in the fashion AI field \cite{rostamzadeh2018fashion, Han_2017_ICCV, Wu_2021_CVPR} often focus on the multimodal retrieval of pre-manufactured products. 
In contrast, the NAIL-STAR benchmark incorporates nail design images that are largely composed of both: (i) a painted part allowing for creative flexibility and (ii) a decorative part that can be modified only by selecting and arranging pre-manufactured embellishments. 
Thus, the NAIL-STAR benchmark requires the models to handle designs that are more diverse than those of existing fashion benchmarks. 
Moreover, most existing benchmarks feature relatively simple and superficial comprehension of multimodal contents.
Most nail image benchmarks do not include a broad spectrum of realistic designs, nor do they provide multi-layered descriptions of the underlying intent of the designs.

\begin{figure*}[t]
    \centering
    \includegraphics[width=\linewidth]{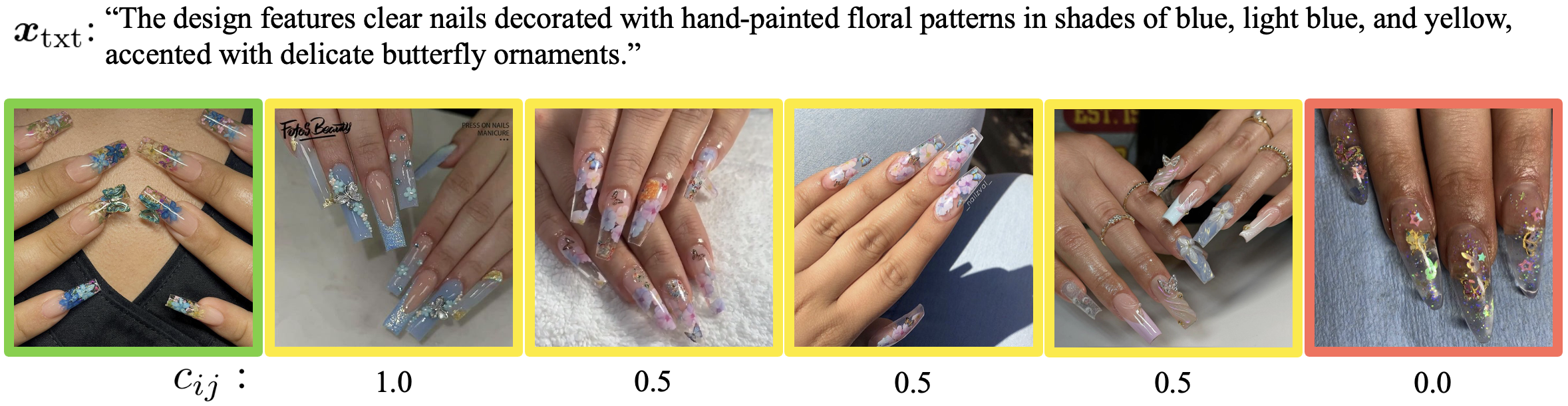}
    \caption{Examples of $c_{ij}$ estimated by the MLLM in CRAM. Positive, unlabeled positive, and negative labels are framed in green, yellow, and red, respectively, with unlabeled positives identified based on $c_{ij}$.}
    \label{fig:dmdl_output}
    \vspace{-2mm}
\end{figure*}

\section{Prompt Examples}
\label{sec:prompt}
\paragraph{IPFM.}
The prompt used in Intent-Palette Fusion Module (IPFM) to obtain $\bm{x}_\text{MDD}$ from $\bm{x}_\text{txt}$ was as follows:
\begin{lstlisting}[breaklines=true,breakindent=0pt, breakautoindent=false,frame=single,basicstyle=\ttfamily\footnotesize]
Extract the design, shape, theme, and impression details from the following nail design request. Only include details that are directly stated or clearly described in the text. Do not infer or mention missing parts.

- Design: Visual details like colors, patterns, textures, or nail accessories.
- Shape: Nail shape or length.
- Theme: The theme, motif, or symbol that represents the design.
- Impression: The impression conveyed by the design or theme.

If any part (design, shape, theme, impression) is not explicitly mentioned, exclude it from the output. There is no need to state that it is missing.

Use this format for the output. Exclude any categories that are not explicitly mentioned:
The design of these nails is ....
The shape of these nails is ....
The theme of these nails is ....
The impression of these nails is ....

nail design request: {dense intent description}
\end{lstlisting}
We replaced the above-mentioned placeholder in {\footnotesize\texttt{\{dense intent description}\}} with $\bm{x}_\text{txt}$.
Similarly, the prompt template to obtain $\bm{x}_\text{NNP}$ from $\bm{x}_\text{txt}$ was defined as follows:
\begin{lstlisting}[breaklines=true,breakindent=0pt, breakautoindent=false,frame=single,basicstyle=\ttfamily\footnotesize]
The following text is a request for a nail design to a nail artist. From this request sentence, create a noun phrase that describes the design, theme, or impression of the nail designs. The sentence can be long, but it should retain all the information and be presented as a single noun phrase. Respond with only the noun phrase.

nail design request: {dense intent description}
\end{lstlisting}
As an example, suppose $\bm{x}_\text{txt}$ was as follows:
\textit{``I'd like a design that primarily features purple, inspired by a wizard, to create a mysterious atmosphere. Please add decorations resembling a ruby on the ring finger. I want long nails with dark marble patterns to enhance the mysterious vibe.''}
For this example, the following output for $\bm{x}_\text{MDD}$ was obtained:
\begin{lstlisting}[breaklines=true,breakindent=0pt, breakautoindent=false,frame=single,basicstyle=\ttfamily\footnotesize]
The design of these nails is primarily purple with dark marble patterns and decorations resembling a ruby on the ring finger. The shape of these nails is long. The theme of these nails is inspired by a wizard. The impression of these nails is mysterious.
\end{lstlisting}
Similarly, $\bm{x}_\text{NNP}$ was obtained as follows:
\begin{lstlisting}[breaklines=true,breakindent=0pt, breakautoindent=false,frame=single,basicstyle=\ttfamily\footnotesize]
A design primarily featuring purple inspired by a wizard with a mysterious atmosphere, long nails with dark marble patterns, and ruby-like decorations on the ring finger.
\end{lstlisting}

\paragraph{VDFM.}
The following prompt was used in Visual Design Fusion Module (VDFM) to generate explanations based on $\bm{x}_\text{img}$ using MLLMs:
\begin{lstlisting}[breaklines=true,breakindent=0pt, breakautoindent=false,frame=single,basicstyle=\ttfamily\footnotesize]
Please describe the nail designs in the image by highlighting the important features, such as colors, patterns, decorations, textures, nail shapes, nail lengths, motifs, or impressions. There's no need to cover everything--focus only on the key distinctive elements.
\end{lstlisting}
Suppose the target nail design image in \cref{fig:qualitative} (a) is given as input.
The following outputs were generated depending on the MLLM used.
GPT-4o generated the following output:
\begin{lstlisting}[breaklines=true,breakindent=0pt, breakautoindent=false,frame=single,basicstyle=\ttfamily\footnotesize]
The nails are long with a coffin shape, featuring a rainbow gradient of bright colors including pink, blue, green, orange, and yellow. Notably, each hand has a focal point decoration: a large, multicolored 3D flower with a smiling face at the center, adding a playful and bold touch. The designs are glossy, which enhances their vibrant appearance.
\end{lstlisting}
In contrast, Qwen2-VL generated:
\begin{lstlisting}[breaklines=true,breakindent=0pt, breakautoindent=false,frame=single,basicstyle=\ttfamily\footnotesize]
The nails in the image feature a vibrant and playful design with the following key elements:\n\n1. **Colors**: The nails are painted in a gradient of bright, neon colors, including pink, blue, green, yellow, and orange. The gradient transitions smoothly from one color to the next.\n\n2. **Patterns**: The nails have a striped pattern with alternating bands of the same colors as the gradient. This creates a cohesive and visually appealing design.\n\n3. **Decorations**: Each nail is adorned with a colorful, three-dimensional flower decoration. The flowers are made of small, round beads in various colors, including red, yellow, green, blue, and purple. Each flower has a smiling face in the center, adding a cheerful and whimsical touch.\n\n4. **Textures**: The nails appear to have a glossy finish, giving them a shiny and reflective look.\n\n5. **Nail Shapes**: The nails are long and coffin-shaped, which is a popular style known as \"stiletto\" nails.\n\n6. **Lengths**: The nails are quite long, extending beyond the natural nail bed, which is typical for stiletto nails.\n\nOverall, the nails combine bright colors, playful patterns, and cute decorations to create a fun and eye-catching design.
\end{lstlisting}

\vspace{-3mm}
\paragraph{CRAM.}
We estimated $c_{ij}$ in Confidence-based Relaxed Alignment Module (CRAM) using MLLMs.
The MLLM output was either ``0,'' ``5,'' or ``10,'' which were scaled to 0.0, 0.5, or 1.0, respectively, and used as $c_{ij}$.
The following prompt was used to estimate $c_{ij}$, which represents the degree to which $\bm{x}_{\text{img}}^{(j)}$ can be considered as an unlabeled positive for $\bm{x}_{\text{txt}}^{(i)}$:
\begin{lstlisting}[breaklines=true,breakindent=0pt, breakautoindent=false,frame=single,basicstyle=\ttfamily\footnotesize]
Your task is to evaluate how well the nails shown in the image match the provided description. Analyze these aspects: colors, patterns, textures, decorations, motifs, nail lengths, nail shapes, and overall impressions. The image is annotated as {candidate noun phrase}. Can it also be described as {gt noun phrase}?
Provide a score of 0, 5, or 10 based on the following criteria:
10: A flawless match with no visible differences or inconsistencies in any aspect.
 Use this score only if the image perfectly aligns with {gt noun phrase}, leaving no room for doubt.
5: The image is very close to being perfectly described as {gt noun phrase}, with only minor and barely noticeable differences
 limited to slight variations in color or nail shape. Differences in illustrations, motifs, or other prominent aspects should result in a score of 0.
0: Use this score even if most aspects are similar, but there is at least one significant and noticeable difference
 that prevents the image from being described as {gt noun phrase}.
State the score (0, 5, or 10) first, followed by a concise reason.
\end{lstlisting}
The above-mentioned placeholders {\footnotesize\texttt{\{candidate noun phrase}\}} and {\footnotesize\texttt{\{gt noun phrase}\}} were replaced with $\bm{x}_{\text{NNP}}^{(j)}$ and $\bm{x}_{\text{NNP}}^{(i)}$, respectively.   
\cref{fig:dmdl_output} shows the resulting scores.
Images that are closely aligned with $\bm{x}_\text{txt}$ are assigned higher scores, whereas those that only partially match $\bm{x}_\text{txt}$ receive lower scores.
In the experiment, MLLM outputs of ``10'' or ``5'' for $\bm{x}_{\text{img}}^{(j)}$ were classified as unlabeled positives for $\bm{x}_{\text{txt}}^{(i)}$ because we set $\theta = 0.5$.

\section{Accelerating Confidence Score Estimation
}
\label{sec:confidence}
In CRAM, we introduce a preprocessing method that speeds up confidence score estimation by filtering candidate pairs according to similarity.
Estimating the scores for all possible pairs in the training set requires performing $(\text{number of descriptions})\times(\text{number of images})$ estimations, resulting in a significant computational cost.
Therefore, it is desirable to perform the estimation process on only a subset of samples that have a high probability of being unlabeled positives.
\textcolor{blue}{
To achieve this, we first obtain the language feature and the visual feature from $\bm{x}_{\text{txt}}^{(i)}$ and $\bm{x}_{\text{img}}^{(j)}$, respectively, using a vision--language model (e.g., CLIP~\cite{radford2021learning}, BEiT-3~\cite{Wang_2023_CVPR}).
We then compute their similarities using cosine similarity~\cite{radford2021learning}.}
Because an image $\bm{x}_{\text{img}}^{(j)}$ with high similarity to $\bm{x}_{\text{txt}}^{(i)}$ is likely an unlabeled positive, we select the top $N_\text{cand}$ images $\left\{ \bm{x}_{\mathrm{img}}^{(k)} \right\}$ (where $k$ represents each element in the index set of these top $N_\text{cand}$ images) with the highest similarity scores as candidate unlabeled positives.

\begin{figure*}[t]
    \centering
    \includegraphics[width=\linewidth]{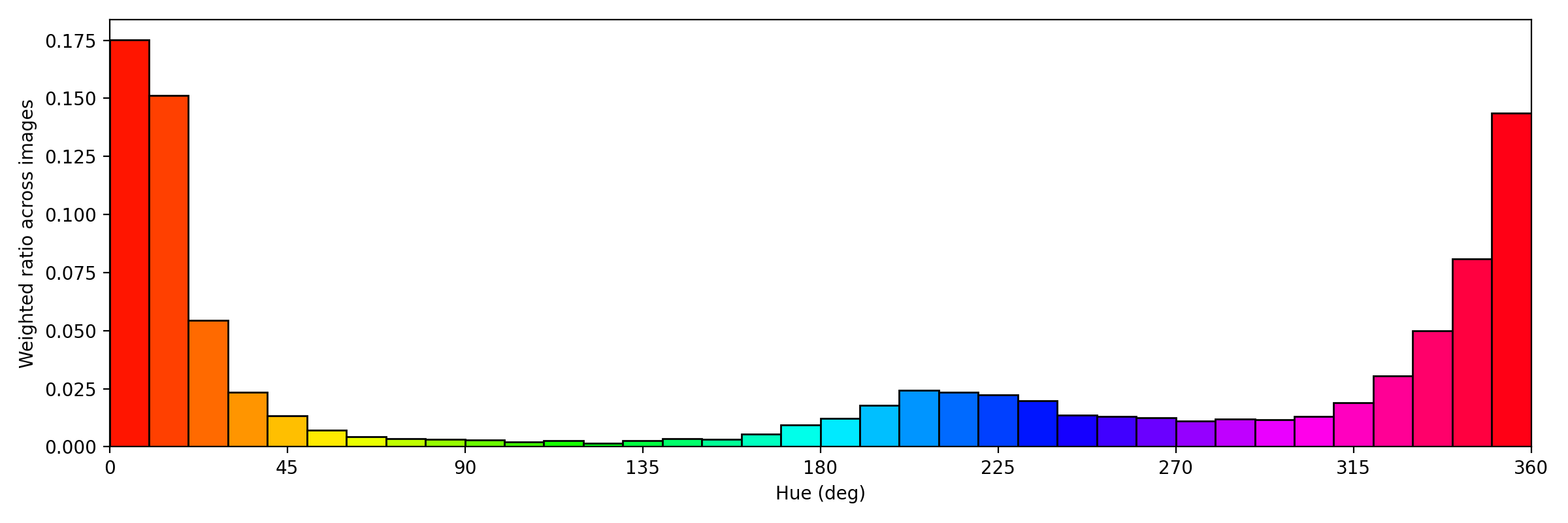}
    \caption{Hue histogram of nail design images. We compute the distribution of hue values (HSV space, 0–360°) using all pixels within segmented nail regions in the NAIL-STAR benchmark. To visualize hue independent of saturation and value, the color of each bar corresponds to the bin center with fixed parameters ($S=1, V=1$). The vertical axis indicates the pixel proportion relative to the total segmented area across the dataset.}
    \label{fig:color_distribution}
\end{figure*}

\vspace{-1mm}
\section{NAIL-STAR Benchmark}
\label{sec:nailster}
\vspace{-1mm}
%5.4
\subsection{Image Filtering}
The collected nail images were pre-processed using the following filtering steps.
(1) Images that showed toenail designs and nail tips were excluded because this study focuses solely on fingernail designs. 
% For this purpose, a filtering method based on CLIP was implemented, leveraging the similarity between each image and the text provided below.
\textcolor{blue}{For this purpose, we employed CLIP-based filtering, an established method~\cite{10376825, dunlap2023diversify} for image selection in dataset construction, leveraging the similarity between each image and the text provided below.}
Specifically, images were removed if they met either of the following criteria: (a) the similarity to ``a photo of fingers'' or ``a photo of nails'' fell below a predetermined threshold, or (b) the similarity to ``a photo of hands'' was lower than that to ``a photo of feet.''

(2) Images in which the nail area was disproportionately small relative to the overall image were excluded.
This is because it is difficult to recognize nail designs in images where the nail area appears small because of the distance from the camera, or where much of the nail is obscured by the angle of the hand.
(3) Images where two or fewer nails were visible were excluded as this made it difficult to grasp the overall design.
(4) Images featuring 11 or more nails were removed because they were presumed to include more than one person's hands, potentially containing a mixture of different nail designs.
As a preprocessing step for the filtering, nail segmentation masks were generated for the images using a segmentation model for fingernails~\cite{nakhoon_dataset}. 
Then, criteria (2) to (4) were applied based on the area and number of nails identified.

(5) To eliminate duplicate images, the Hamming distance between the hash values of each image was calculated, and identical or extremely similar images were filtered out.
(6) A manual inspection was conducted to exclude any remaining images deemed unsuitable for this study, such as AI-generated images or those with extremely poor lighting that hindered the recognition of nail designs.

\cref{fig:color_distribution} shows the distribution of hue values across all segmented nail regions in the NAIL-STAR benchmark. We extracted pixel values in the HSV color space and constructed a histogram discretized over the range $0^\circ \text{--} 360^\circ$.
The high proportion of warm colors is attributable not only to a bias toward warm-colored nail designs, but also to the warm hue of the underlying nail plate visible through clear gel.

%5.2
\subsection{Annotation of Descriptions}
\begin{figure}[t]
    \centering
    \includegraphics[width=\linewidth]{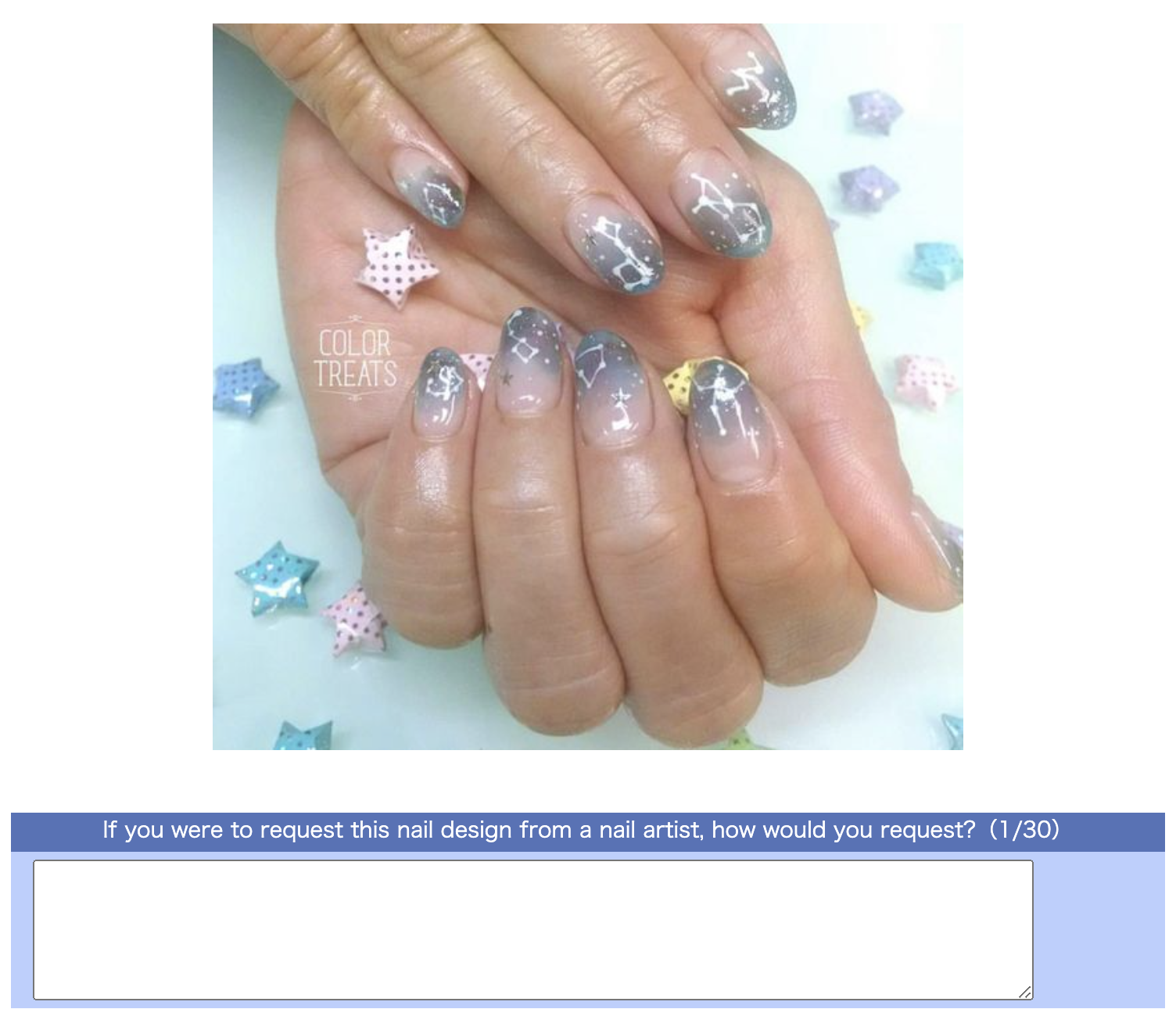}
    \caption{Description annotation interface. The annotators were requested to input a dense intent description for the displayed nail design image into the text box below the image.}
    \label{fig:annotation}
\end{figure}

We annotated dense intent descriptions for nail design images as described below.
\cref{fig:annotation} shows the annotation interface. 
The annotators were presented with a series of nail design images, one at a time, and were instructed to respond to the following question: \textit{``If you were to request this nail design from a nail artist, how would you request it?''}
They were instructed to provide a visual description of each nail design, encompassing elements such as color, pattern, texture, and nail embellishments. 
Moreover, they were asked to include, where relevant, any themes suggested by the visual characteristics and the overall impression conveyed by the design.
To avoid simple image captions, the annotators were instead instructed to imagine a scenario in which they were describing a nail design to a nail artist and to express their intent in a detailed and multi-layered manner. 

\textcolor{blue}{We implemented a multi-step quality control process for the reliability of the crowdsourced annotations. First, we randomly sampled and manually inspected five annotations from each annotator. All data from annotators with systematic errors were excluded. Second, we employed the Polos score~\cite{10656819}, an automatic evaluation metric for image captions, to identify potentially low-quality annotations. All samples with a Polos score below 0.2 were manually reviewed, and erroneous annotations were removed. Finally, every sample in the test set was manually checked.}

\begin{figure*}[t]
    \centering
    \includegraphics[width=\linewidth]{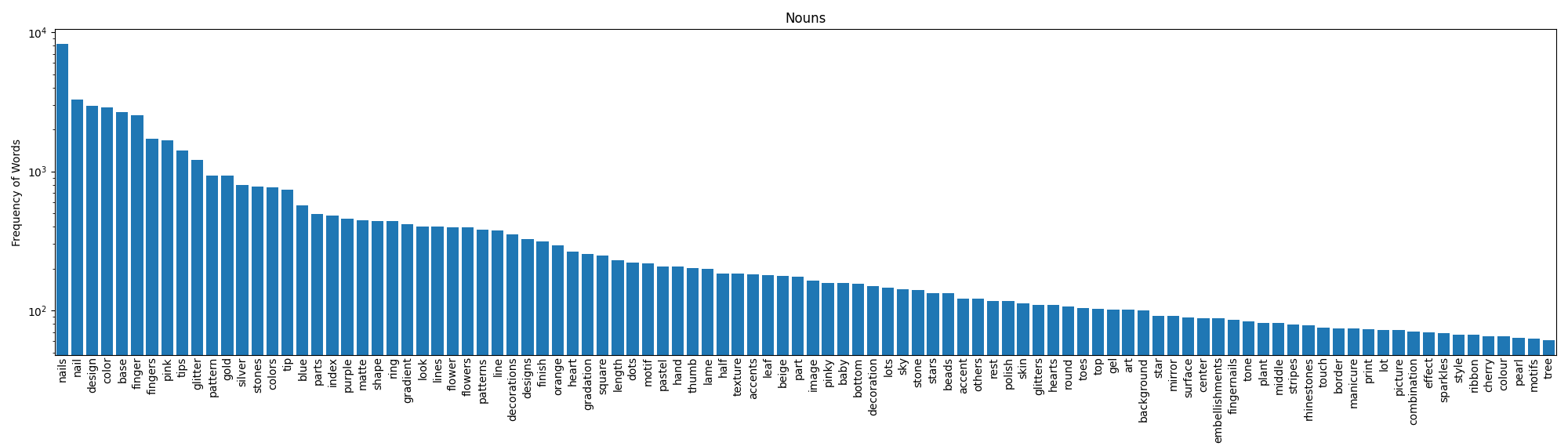}
    \includegraphics[width=\linewidth]{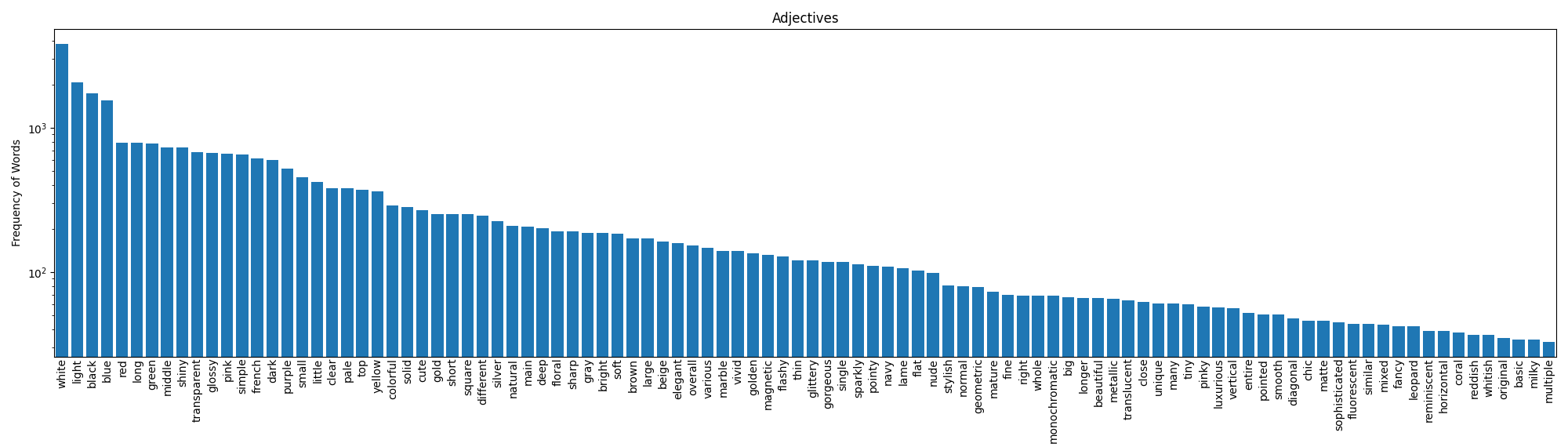}
    \includegraphics[width=\linewidth]{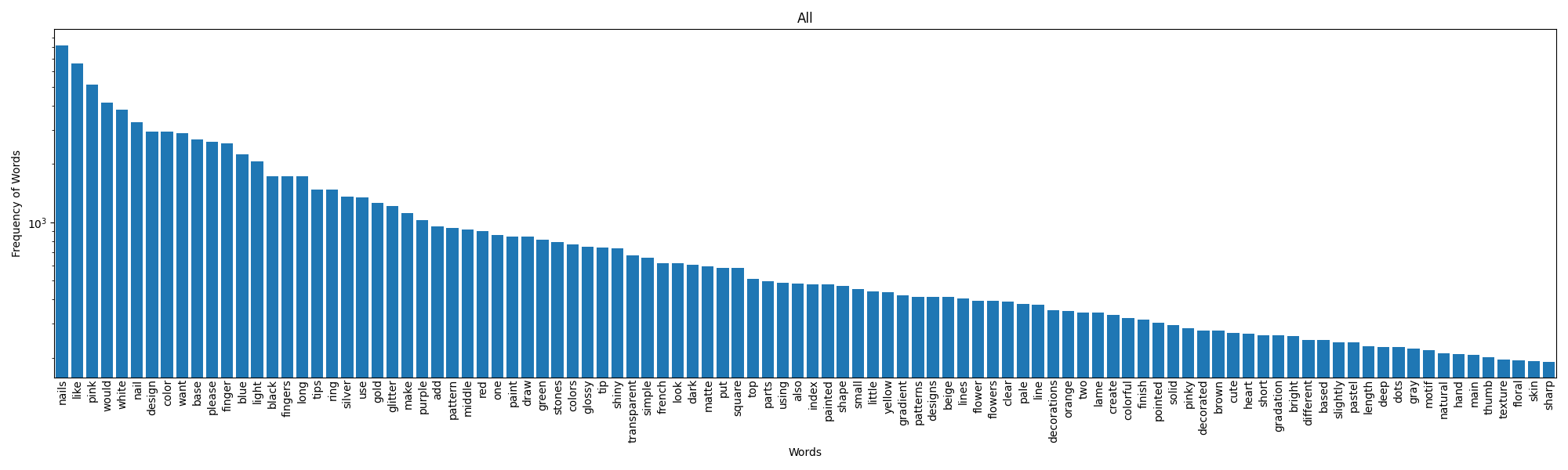}
    \caption{The word frequency distribution of the NAIL-STAR benchmark. The frequency distributions of the 100 most common nouns, adjectives, and all words are presented. Singular and plural forms of nouns, comparative and superlative forms of adjectives, and present and past participles of verbs were counted independently. Stopwords were removed.}
    \vspace{-4mm}
    \label{fig:word_frequencies}
\end{figure*}

Data collection was facilitated through the use of SoSci Survey.
All descriptions were required to be in English and at least 10 words.
A total of 10,625 descriptions were collected from 208 annotators, with an average of 51.1 descriptions per annotator.
\cref{fig:word_frequencies} shows the word frequency distribution of the annotations.

\subsection{Annotation of Palette Queries}
\label{sec:palette-annotation}
We automatically annotated palette queries for the training and validation sets of the NAIL-STAR and Marqo Fashion200K benchmarks as follows.
First, we generated segmentation masks from $X_{\text{img}}$ using a fingernail segmentation model~\cite{nakhoon_dataset} for the NAIL-STAR benchmark and a segmentation model~\cite{Xie2021SegFormerSA} fine-tuned on a fashion dataset for the Marqo Fashion200K.
Within each segmented region, we applied SLIC to partition the area into $N_\text{SLIC}$ superpixels, where $N_\text{SLIC} \ge 1000$. 
For each superpixel, we computed the mean RGB value and pixel count, defining this mean as the superpixel's representative color.
These colors were subsequently converted to the CIELAB color space to enable perceptually meaningful comparisons.

Next, we computed a pairwise distance matrix between representative colors in the CIELAB space using the CIEDE2000 color difference $\Delta E_{00}$ and performed agglomerative clustering with average linkage. 
Clusters were iteratively merged until the average inter-cluster distance exceeded a threshold $\theta_{\text{cluster}}$. 
The representative color of each cluster was defined as the color of the superpixel that minimized the weighted sum of $\Delta E_{00}$ distances to all other superpixels in the same cluster, and its CIELAB value was converted back to RGB.

Finally, we sorted the clusters by area ratio in descending order to select the palette query.
Iterating from the largest to the smallest cluster, we retained a cluster as a color of the palette query only if its individual area ratio was at least $\theta_{\text{min}}$ and the cumulative area ratio of selected clusters did not exceed $\theta_{\text{cum}}$. 

\begin{figure}[t]
    \centering
    \includegraphics[width=0.95\linewidth]{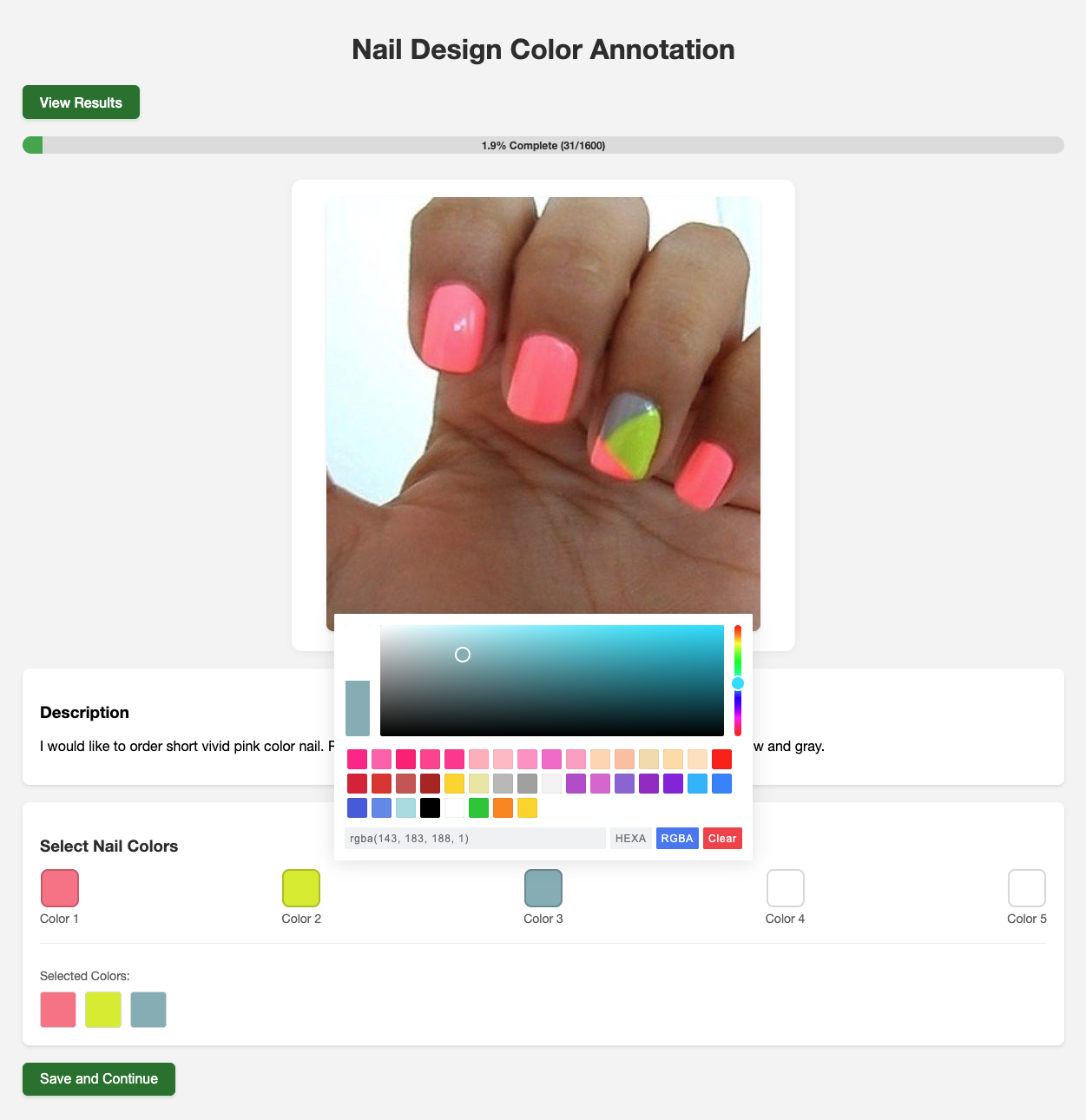}
    \vspace{-1mm}
    \caption{Palette query annotation interface. The annotators were requested to specify a palette query for the displayed nail design by using a color picker, selecting any number of colors (0–5).}
    \label{fig:annotation_color}
    \vspace{-4mm}
\end{figure}

In contrast, we manually annotated palette queries for the test sets of both benchmarks as follows.
As illustrated in \cref{fig:annotation_color}, annotators were presented with both the target image and its corresponding description, utilizing an interface that mimicked a typical e-commerce search UI with a color picker.
Annotators were allowed to select between zero and five colors for each sample.
Rather than exhaustively specifying all colors present in the image, they were instructed to choose only the key colors they would input if they were actually searching for that design.
For efficiency, we provided a set of representative colors as initial candidates. 
However, annotators were not restricted to these presets and could freely adjust the color picker to any arbitrary color before finalizing the query.

\section{Implementation Details}
\vspace{-1mm}
\label{sec:implementation}
%5.10
We trained our model on a GeForce RTX 4090 with 24 GB of VRAM and an Intel Core i9-13900KF with 64 GB of RAM.
%5.11
% The training time for the proposed model was approximately 20 minutes, and the inference time for computing the similarity between a single description and 100 nail design images was approximately 0.37 seconds.
The training time for the proposed model was approximately 20 minutes.
All features and generated outputs from LLMs and MLLMs required for training were obtained offline prior to the training process, because these models remain frozen and do not need to be recomputed during training.
The average total computational time per sample for feature extraction and generation by LLMs and MLLMs was approximately 4.2 seconds.
At inference time, the similarity between a single description and 1,600 nail design images was computed in approximately 0.8 seconds, where the features from $\bm{x}_{\text{txt}}$ were obtained online, while those from $\bm{x}_{\text{img}}$ were pre-extracted.
\textcolor{blue}{In practical text-to-image retrieval, the candidate pool of images is predetermined. Therefore, it is feasible to obtain and store the visual features in advance, thereby significantly reducing the computational burden during inference.
Moreover, this latency of 0.8 seconds is considered acceptable, as users typically do not perceive delays under 1 second as problematic~\cite{10.1145/2600428.2609627}.}

%5.8
We used the Adam optimizer with $\beta_{1} = 0.9$, $\beta_{2} = 0.98$, and a fixed learning rate of $1 \times 10^{-5}$.
We set the batch size to 64 and trained the model for 40 epochs.
The number of candidate images was set to $N_\text{cand} = 30$.
The weights of the loss functions were set to $\lambda_\text{UP} = 0.7$, $\lambda_\text{N} = 0.7$.
The language representations $\left( \bm{l}_\text{txt}, \bm{l}_\text{MDD}, \bm{l}_\text{NNP} \right)$, the palette representation $\bm{p}$, and the language–palette representation $\bm{l}_\text{+}$ described in~\Cref{sec:ipfm}, as well as the visual representations $\bm{v}_{\text{s}}^{(i)}$, $\bm{v}_{\text{a}}^{(i)}$, $\bm{v}_{\text{n}}^{(i)}$, and $\bm{v}$ detailed in~\Cref{sec:vdfm}, are all $1024$-dimensional.
%5.13
In the CRAM, the MLLM output was either ``0,'' ``5,'' or ``10,'' which were scaled to 0.0, 0.5, or 1.0, respectively, and used as $c_{ij}$.
%5.9
Our model had approximately $4.3 \times 10^7$ trainable parameters and $1.2 \times 10^9$ multiply-add operations.
%5.12
We calculated the recall@1 on the validation set for each epoch. 
For the evaluation on the test set, we used the model with the maximum recall@1 on the validation set.

\section{Discussion on Token Length Limitations}
\label{sec:token_length_limitation}
In our experiments, all models were fine-tuned on the NAIL-STAR and Marqo Fashion200K benchmarks, where descriptions exceeding the maximum token length of each text encoder were truncated.
However, the token length limitations of text encoders did not significantly affect the performance of the methods in the comparisons.
Among all baselines, SigLIP~\cite{10377550} has the shortest maximum input length, limited to 64 tokens.
Our method employed BEiT-3~\cite{Wang_2023_CVPR} and SigLIP as text encoders.
To ensure a fair comparison, the BEiT-3 text encoder was consistently configured with a maximum input length of 64 tokens in both the baseline and our proposed method.
The other baseline methods were evaluated using the default maximum token lengths defined in their official implementations (e.g., 77 tokens for CLIP~\cite{radford2021learning}).
Therefore, the token length limitations did not provide any advantage to our method.
Furthermore, in the NAIL-STAR benchmark, only 0.42\% of the descriptions exceed 64 tokens, suggesting that the impact of information loss due to truncation on the evaluation results was minimal.

\section{Evaluation Metrics}
%6.4
We used the mean reciprocal rank (MRR) and recall@$K (K=1,10)$ as evaluation metrics, with recall@1 as the primary metric.
The MRR is defined as follows:
\begin{align}
\text{MRR} = \frac{1}{N_{\text{txt}}} \sum_{i=1}^{N_{\text{txt}}} \frac{1}{r_1^{(i)}},
\end{align}
where $N_{\text{txt}}$ and ${r_1^{(i)}}$ denote the number of descriptions and rank of the target nail design image for the $i$-th description, respectively.
Recall@$K$ is defined as follows:
\begin{align}
\text{Recall@}K = \frac{1}{N_{\text{txt}}} \sum_{i=1}^{N_{\text{txt}}} \frac{|A_i \cap B_i|}{|A_i|},
\end{align}
where $A_i$ and $B_i$ denote the set of nail design images to be retrieved for the $i$-th description and the set of top-$K$ images for the $i$-th description, respectively.

\section{Additional Quantitative Results}

\begin{table}[!t]
    \centering
    \small
    \resizebox{\columnwidth}{!}{
    \begin{tabular}{lcccc}
        \toprule
        Method & MRR↑ [\%] & R@1↑ [\%] & R@10↑ [\%] \\
        \hline
        CLIP~\cite{radford2021learning} & 12.4 & 6.5 & 23.7 \\
        FashionViL~\cite{han2022fashionvil} & 2.3 & 0.6 & 4.8 \\
        FAME-ViL~\cite{Han_2023_CVPR} & 20.2 & 11.2 & 39.3 \\
        BEiT-3~\cite{Wang_2023_CVPR} & 34.9 & 23.5 & 57.9 \\
        BLIP-2~\cite{pmlr-v202-li23q} & 14.4 & 7.5 & 28.0 \\
        SigLIP~\cite{10377550} & 31.4 & 21.3 & 51.6 \\
        Alpha-CLIP~\cite{Sun_2024_CVPR} & 19.6 & 12.1 & 34.3 \\
        Long-CLIP~\cite{zhang2025long} & 10.6 & 5.6 & 19.7 \\
        MM-Embed~\cite{lin2025mmembed} & 15.3 & 9.1 & 27.8 \\
        LamRA~\cite{liu2025lamra} &  15.0 & 9.9 & 28.6 \\
        \bottomrule
    \end{tabular}
    }
    \vspace{-1mm}
    \caption{\textcolor{blue}{Quantitative results of baseline methods on the NAIL-STAR benchmark in the zero-shot settings. The scores are reported based on a single trial.}}
    \label{tab:zero-shot}
\end{table}

\begin{table}[!t]
    \centering
    \small
    \resizebox{\columnwidth}{!}{
    \begin{tabular}{lcccc}
        \toprule
        Token Length & \# & MRR↑ [\%] & R@1↑ [\%] & R@10↑ [\%] \\
        \hline
        0--19  & 781  & 57.9 & 45.3 & 82.5 \\
        20--29 & 560  & 60.0 & 48.0 & 81.4 \\
        30--39 & 187  & 68.4 & 58.3 & 89.3 \\
        40--69 & 72   & 78.6 & 69.4 & 94.4 \\
        \hline
        Total  & 1600 & 60.8 & 48.9 & 83.4 \\
        \bottomrule
    \end{tabular}
    }
    \vspace{-1mm}
    \caption{\textcolor{blue}{Retrieval performance of NaiLIA across subsets grouped by description token length on the NAIL-STAR benchmark, evaluated in the description-only setting without $\bm{x}_\text{pal}$. The scores are reported based on a single trial.}}
    \label{tab:token-length}
\end{table}

\subsection{Zero-Shot Evaluation}
\label{sec:zero-shot}
\textcolor{blue}{\Cref{tab:zero-shot} shows the results of the zero-shot evaluation of the baseline models on the NAIL-STAR benchmark.}
\textcolor{blue}{The scores are reported based on a single trial, since these provide consistent results across multiple trials. Even for the best-performing method BEiT-3, the recall@1 score remains at only 23.5\%. This result highlights the difficulty of this task.}

\subsection{Effect of Query Length on Performance}
\textcolor{blue}{
To analyze how performance varies with query length, we conducted additional evaluations. 
We divided the test set into subsets according to the token length of the descriptions and evaluated NaiLIA in the description-only setting, where $\bm{x}_\text{pal}$ is not provided.
\Cref{tab:token-length} shows the distribution of token lengths and the evaluation results for each subset. 
The results indicate that MRR and recall@1 scores increase as the token length grows. 
These findings demonstrate that NaiLIA is robust across all token lengths and is particularly effective for longer descriptions containing dense intent.
}

\begin{table}[!t]
    \centering
    \small
    \begin{tabular}{lcccc}
        \toprule
        Model & MRR↑ [\%] & R@1↑ [\%] & R@10↑ [\%] \\
        \hline
        (a) NaiLIA (full) & 82.5 & 74.5 & 96.6 \\
        (b) w/o $\bm{v}_\text{s}$ & 81.6 & 73.3 & 95.7 \\
        (c) w/o $\bm{v}_\text{a}$ & 41.1 & 28.5 & 65.7 \\
        (d) w/o $\bm{v}_\text{n}$ & 78.8 & 69.7 & 95.1 \\
        \bottomrule
    \end{tabular}
    \vspace{-1mm}
    \caption{\textcolor{blue}{Ablation study of VDFM on the  Marqo Fashion200K. The scores are reported based on a single trial.}}
    \label{tab:ablation_fashion_vdfm}
\end{table}

\begin{figure}[t]
    \centering
    \includegraphics[width=\linewidth]{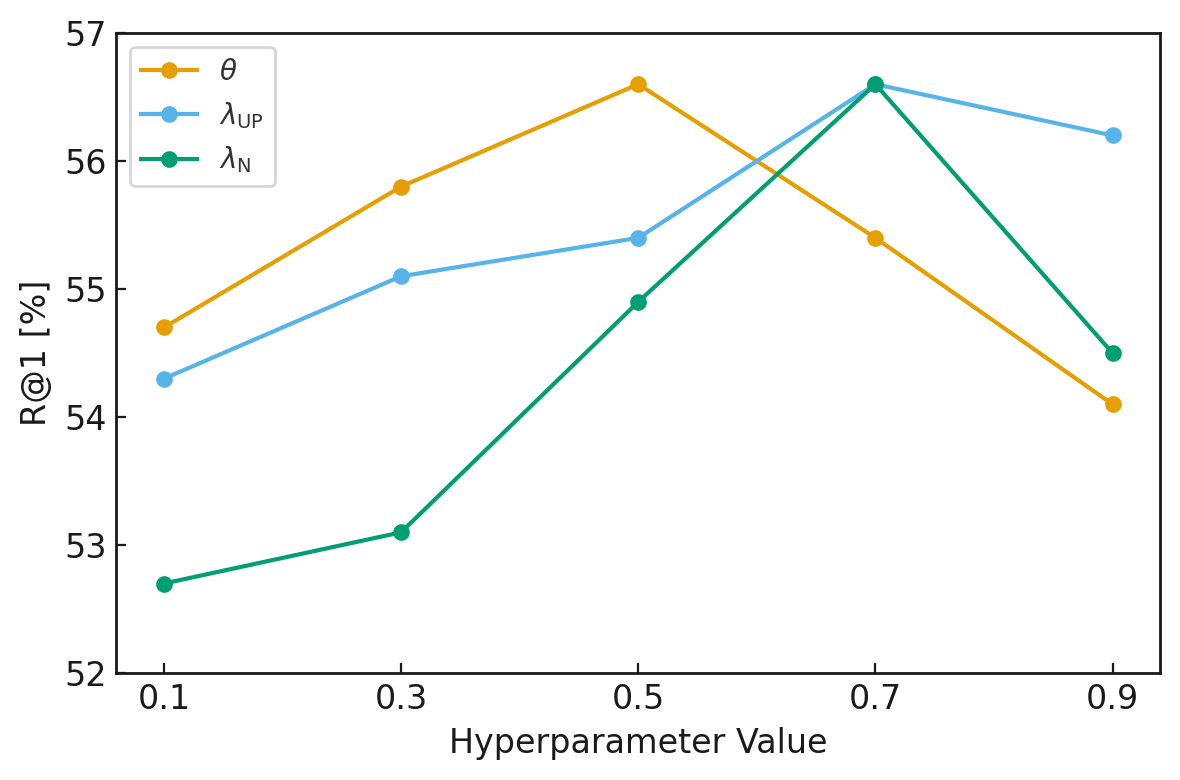}
    \caption{Sensitivity analysis of the CRC loss with respect to the confidence threshold $\theta$ and the balance terms $\lambda_{\text{UP}}$ and $\lambda_{\text{N}}$. For each curve, we varied one hyperparameter over $\{0.1, 0.3, 0.5, 0.7, 0.9\}$ while keeping the other two fixed at their default values ($\theta = 0.5$, $\lambda_{\text{UP}} = 0.7$, $\lambda_{\text{N}} = 0.7$). The scores are reported based on a single trial.}
    \label{fig:sensitivity}
\end{figure}

\subsection{Ablation Study of VDFM}
\label{sec:vdfm_ablation}
\textcolor{blue}{To demonstrate the generality of VDFM, we conducted an ablation study on the Marqo Fashion200K.
\Cref{tab:ablation_fashion_vdfm} shows that recall@1 drops by 1.2, 46.0, and 4.8 points for Models (b), (c), and (d), respectively, compared with Model (a).
This suggests that the VDFM is impactful across domains and not limited to a specific setting.
}

\begin{table*}[!ht]
    \centering
    \small
    \begin{tabular}{clcc}
        \toprule
        Errors Type & Description & \#Error (desc-only) & \#Error (full) \\
        \hline
        \multirow{5}{*}[0pt]{\centering Comprehension} 
        & Comprehension errors in patterns and decorative elements & \textbf{19} & \textbf{20} \\
        & Comprehension errors in design themes & 17 & 19 \\
        & Comprehension errors in color-related expressions & 13 & 6 \\
        & Comprehension errors in nail shape and decoration placement & 11 & 11 \\
        & Comprehension errors in the correspondence between fingers and design & 5 & 7 \\
        \hline
        \multirow{3}{*}[0pt]{\centering Ambiguity}  
        & Ambiguity in color description & 18 & 9 \\
        & Ambiguity in impression description & 9 & 18 \\
        & Ambiguous descriptions unrelated to color or impression & 8 & 10 \\
        \hline
        & Total & 100 & 100 \\
        \bottomrule
    \end{tabular}
    \caption{Number of failure cases in each category for NaiLIA in the description-only and full settings on the NAIL-STAR benchmark. The largest count in each setting is highlighted in \textbf{bold}.}
    \label{tab:error_analysis_combined}
\end{table*}

\subsection{Sensitivity Analysis}
We conducted a sensitivity analysis to examine how each hyperparameter in the CRC loss influences retrieval performance, as illustrated in Fig.~\ref{fig:sensitivity}.
In this analysis, we varied one hyperparameter over $\{0.1, 0.3, 0.5, 0.7, 0.9\}$ while keeping the remaining two fixed at their default values ($\theta = 0.5$, $\lambda_{\text{UP}} = 0.7$, $\lambda_{\text{N}} = 0.7$).
The scores are reported based on a single trial.
The results demonstrate the robustness of the CRC loss to variations in hyperparameters. Similar trends were also observed across other parameter combinations.

\subsection{MLLM-Based Retrieval}
We employed MLLMs to estimate $c_{ij}$ in CRAM because MLLMs can approximate similarity scores between text and image pairs.
However, retrieval based solely on similarity scores predicted by MLLMs is impractical due to their high computational cost and limited performance.
In practice, computing the similarity between a single description and 1,600 nail design images took an average of 18.7 minutes.
When performing inference on 10 descriptions, the recall@1 was limited to 20.0\% (with the target nail design image ranked 28th on average).
The experiment was conducted using Qwen2-VL~\cite{Qwen2VL} as the MLLM, with a batch size of 128 for batch processing.
The MLLM was prompted to output an integer similarity score from 0 to 9 for each image, and the images were ranked in descending order of these scores.
In cases where multiple images were assigned the same score, their order was randomized within the tied rank.
These results highlight the practical limitations of using MLLMs directly for retrieval.

\section{Error Analysis}
\label{sec:error}
%6.11
To investigate the limitations of the proposed method, we analyzed cases where the method did not perform as expected. 
% We defined descriptions for which the target nail design image was retrieved at a rank lower than 10 as failure.
% There were a total of 416 failure cases within the NAIL-STAR benchmark test set.
%6.12
We analyzed the 100 failure cases with the lowest ranks for the target nail design images.
% \Cref{tab:error_analysis} shows the categorizations of the failure modes.
\Cref{tab:error_analysis_combined} shows the categorizations of the failure modes for both the description-only and full settings of NaiLIA on the NAIL-STAR benchmark.
The failure modes could be grouped into the following eight categories:
\begin{itemize}
    \setlength{\parskip}{0.2mm} % 段落間
    \setlength{\itemsep}{0.2mm} % 項目間
    \item Comprehension errors in patterns and decorative elements: This category encompasses cases where the model incorrectly interpreted patterns or decorative elements in the nail design. This includes cases where these elements were relatively small compared with the overall image or had colors similar to the background, making them difficult to recognize.
    \item Comprehension errors in design themes: This category covers cases where the model failed to comprehend the themes of the nail design, such as those inspired by specific objects or concepts.
    \item Comprehension errors in color-related expressions: This refers to errors where the model wrongly interpreted color-related expressions. For example, there were cases where nail designs with separate pink and gray colors were ranked higher than those with a blended hue, which was misaligned with user intent, even when the description specified ``a color that's between pink and gray.''
    \item Comprehension errors in nail shape and decoration placement: This category includes cases where the nail length, shape, or decoration placement was misinterpreted by the model. For example, there were cases where nail designs with glitter applied at the base rather than the tips were included, even when the description explicitly specified ``glitter on the tips.''
    \item Comprehension errors in the correspondence between fingers and design: This category refers to errors in which the model failed to comprehend the correspondence between the finger and the design. Specifically, this includes cases where images in which the painting or decoration of the nail on a different finger was incorrectly ranked higher than the target nail design image.
    \item Ambiguity in color description: This error arises from the ambiguity inherent in color description. For example, given a description specifying ``pink,'' the model preferentially retrieved nail designs with a reddish-pink hue, even if the user wanted a yellowish one.
    \item Ambiguity in impression description: This category comprises cases where the given description was ambiguous in terms of impression-related expressions, such as ``flashy'' or ``mysterious.''
    \item Ambiguous descriptions unrelated to color or impression: This refers to cases where numerous unlabeled positive samples dominated the top rankings.
\end{itemize}

\section{Discussion on LLM Hallucinations}
\label{sec:discussion_hallucinations}
\textcolor{blue}{We investigated how potential LLM hallucinations or parsing errors might propagate to retrieval quality. Specifically, for the 100 failure cases analyzed in \Cref{sec:error}, we manually examined whether any hallucinations were present in the multi-layered design descriptions and normalized noun phrases corresponding to each description. As a result, none of the 100 failure cases analyzed involved hallucinations caused by the LLM. This can be attributed to the fact that the prompts explicitly instructed the LLM to use only information present in the original text, thereby reducing the likelihood of hallucination. Although hallucinations may exist outside these cases, the integration of three types of features in IPFM appears to have prevented such hallucinations from resulting in critical errors.
}

\section{Additional Qualitative Results}
\label{sec:additonal_qualitative}
\begin{figure*}[t]
    \centering
    \includegraphics[width=\linewidth]{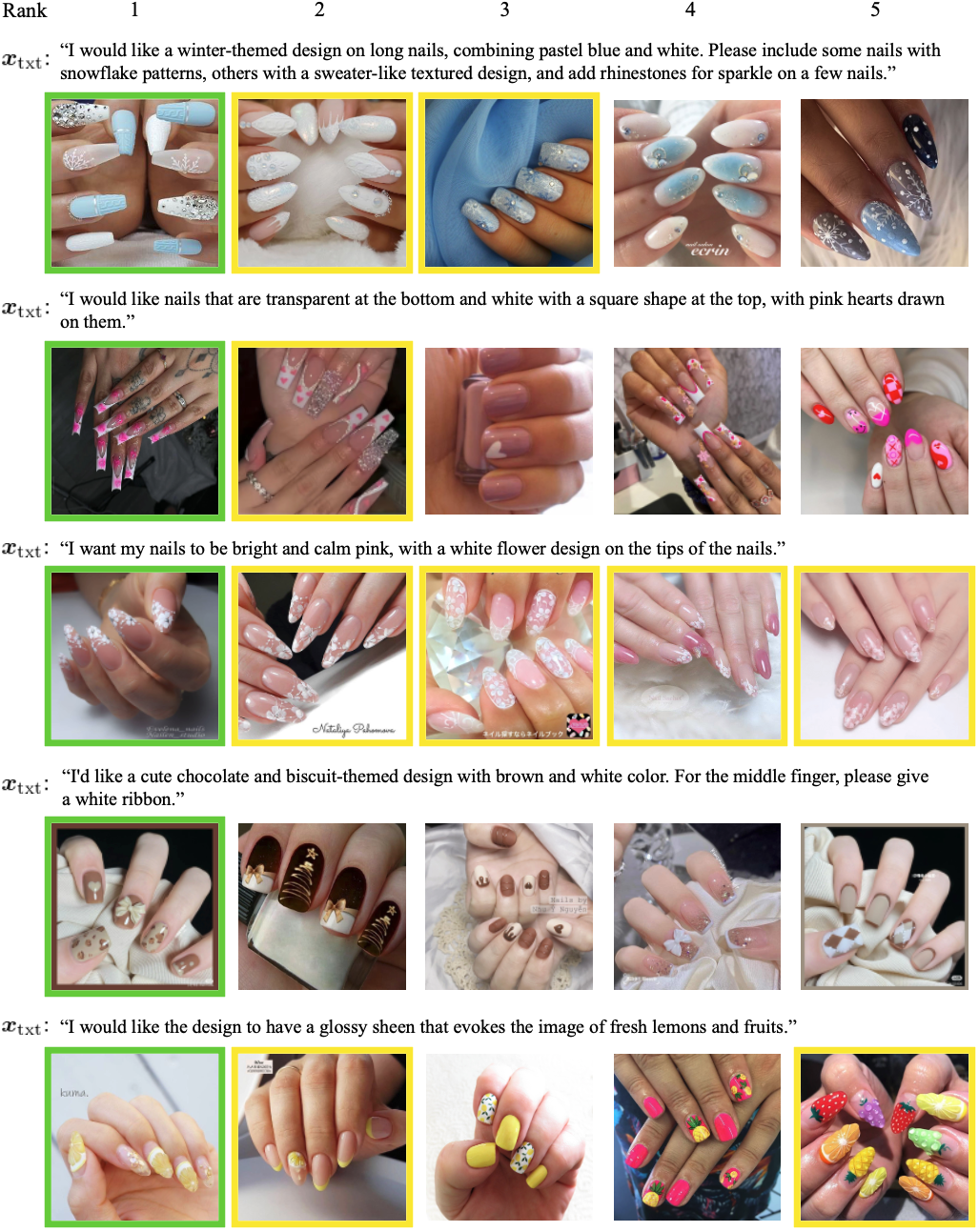}
    \caption{Additional qualitative results of NaiLIA in the \textbf{description-only} setting, where only $\bm{x}_{\text{txt}}$ is provided. The top-5 retrieved images are shown. Positives and unlabeled positives are enclosed in green and yellow frames, respectively.}
    \label{fig:additional_qualitative_1}
\end{figure*}

\begin{figure*}[t]
    \centering
    \includegraphics[width=\linewidth]{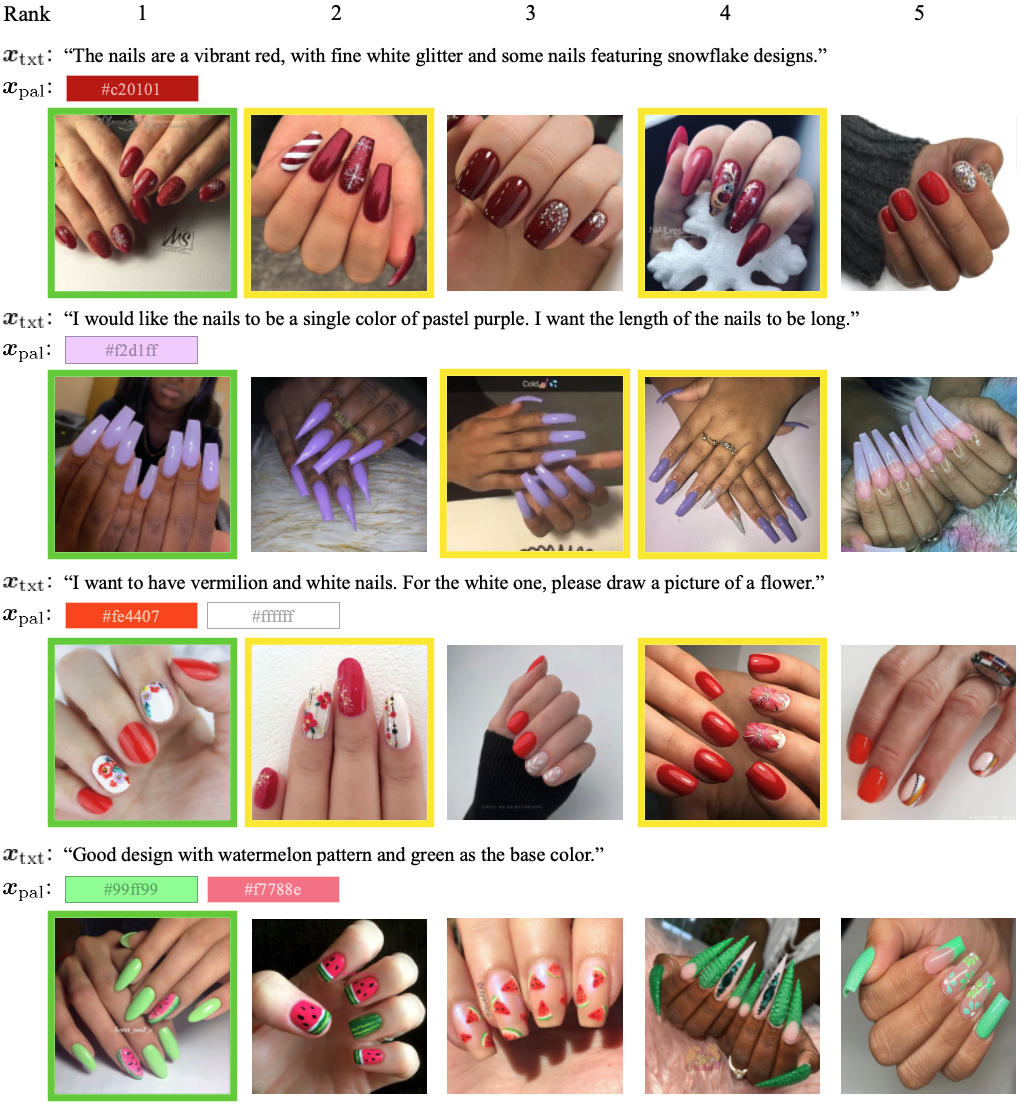}
    \caption{Additional qualitative results of NaiLIA in the setting where \textbf{both} $\bm{x}_{\text{txt}}$ and $\bm{x}_{\text{pal}}$ are provided. The top-5 retrieved images are shown. Positives and unlabeled positives are enclosed in green and yellow frames, respectively.}
    \label{fig:additional_qualitative_2}
\end{figure*}

\begin{figure*}[t]
    \centering
    \includegraphics[width=\linewidth]{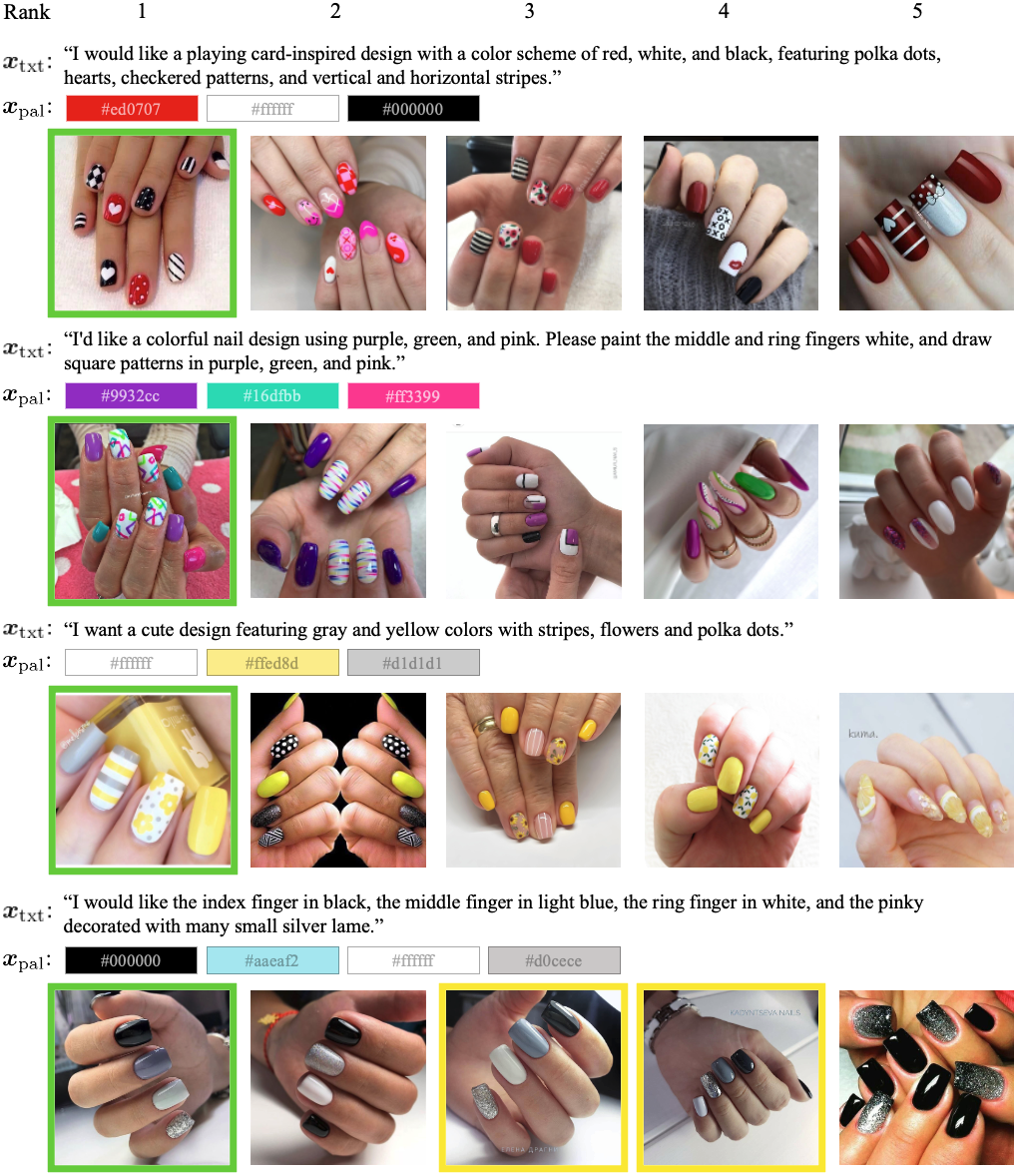}
    \caption{Additional qualitative results of NaiLIA in the setting where \textbf{both} $\bm{x}_{\text{txt}}$ and $\bm{x}_{\text{pal}}$ are provided. The top-5 retrieved images are shown. Positives and unlabeled positives are enclosed in green and yellow frames, respectively.}
    \label{fig:additional_qualitative_3}
\end{figure*}

\cref{fig:additional_qualitative_1} presents additional qualitative results of NaiLIA in the description-only setting, where only $\bm{x}_{\text{txt}}$ is provided. 
\cref{fig:additional_qualitative_2} and \cref{fig:additional_qualitative_3} show 
results of NaiLIA in the setting where both $\bm{x}_{\text{txt}}$ and $\bm{x}_{\text{pal}}$ are provided.
The top-5 retrieved images are shown. 
Positives and unlabeled positives are enclosed in green and yellow frames, respectively.

\footnotetext[1]{"NaiLIA" is a coined name for our proposed method and not an abbreviation (similar to Retrospex~\cite{xiang-etal-2024-retrospex} and BiasX~\cite{zhang-etal-2023-biasx})}

% \section{URLs of Images in This Paper}
% The images included in this paper are available at the URLs listed in \Cref{tab:image_url_1} and \Cref{tab:image_url_2}.

% \clearpage
% \input{tab/image_url_1}
% \input{tab/image_url_2}

\end{document}